\documentclass[10pt]{article}
\usepackage[utf8]{inputenc}
\usepackage[margin=1in]{geometry}
\usepackage{rotating}
\usepackage{booktabs}
\usepackage{array}
\usepackage{amsmath, amsfonts, amssymb}
\usepackage[table,xcdraw]{xcolor}
\usepackage{hyperref}
\usepackage{bbm}
\usepackage{algorithm}
\usepackage[noend]{algpseudocode}
\usepackage{subcaption}
\usepackage{graphbox}
\usepackage{accents}

\usepackage{hhline}
\usepackage{colortbl}

\usepackage{tabularx,booktabs}
\newcolumntype{Y}{>{\centering\arraybackslash}X}

\usepackage{makecell}

\usepackage[numbers,sort]{natbib}

\usepackage{mathtools}
\mathtoolsset{showonlyrefs}

\usepackage{amsthm}
\usepackage[normalem]{ulem}
\usepackage{soul}

\newtheorem{theorem}{Theorem}

\newtheorem{prop}{Proposition}

\newtheorem{assumption}{Assumption}

\newtheorem{example}{Example}

\usepackage{chngcntr}
\usepackage{apptools}
\AtAppendix{\counterwithin{theorem}{section}}
\AtAppendix{\counterwithin{prop}{section}}
\AtAppendix{\counterwithin{cor}{section}}
\AtAppendix{\counterwithin{lemma}{section}}

\usepackage{mathtools}

\usepackage{multicol}
\usepackage{multirow}

\definecolor{Gray}{gray}{0.8}
\definecolor{LightGray}{gray}{0.95}

\def\R{\mathbb{R}}
\def\E{\mathbb{E}}

\newcommand{\X}{\mathcal{X}}

\newcommand{\F}{\mathcal{F}}
\newcommand{\C}{\mathcal{C}}

\renewcommand{\P}{\mathbb{P}}
\newcommand{\GG}{\mathbb{G}}

\newcommand{\cN}{\mathcal{N}}

\newcommand{\piopt}{\pi_{\mathrm{opt}}}

\newcommand{\one}{\mathbf{1}}

\newcommand{\Var}{\mathrm{Var}}

\newcommand{\cd}{\stackrel{d}{\to}}
\newcommand{\cp}{\stackrel{p}{\to}}

\newcommand{\cH}{\mathcal{H}}

\newcommand{\nb}{n_b}
\newcommand{\nlab}{n_{\mathrm{lab}}}
\newcommand{\nlabt}{n_{\mathrm{lab},t}}
\newcommand{\nlabtminone}{n_{\mathrm{lab},t-1}}

\usepackage{esvect}

\DeclareMathOperator*{\argmax}{arg\,max}
\DeclareMathOperator*{\argmin}{arg\,min}

\makeatletter
\def\blfootnote{\xdef\@thefnmark{}\@footnotetext}
\makeatother

\title{Active Statistical Inference}

\author{Tijana Zrnic$^*$\quad \quad Emmanuel J. Cand\`es$^\dagger$\\ \\ 
\texttt{\{tijana.zrnic,candes\}@stanford.edu }\\
$^*$Department of Statistics and Stanford Data Science\\
$^\dagger$Department of Statistics and Department of Mathematics\\ Stanford University}
\date{}

\begin{document}

\maketitle

\begin{abstract}
Inspired by the concept of active learning, we propose \emph{active inference}---a methodology for statistical inference with machine-learning-assisted data collection. Assuming a budget on the number of labels that can be collected, the methodology uses a machine learning model to identify which data points would be most beneficial to label, thus effectively utilizing the budget.
It operates on a simple yet powerful intuition: prioritize the collection of labels for data points where the model exhibits uncertainty, and rely on the model's predictions where it is confident.
Active inference constructs provably valid confidence intervals and hypothesis tests while leveraging any black-box machine learning model and handling any data distribution. The key point is that it achieves the same level of accuracy with far fewer samples than existing baselines relying on non-adaptively-collected data. This means that for the same number of collected  samples, active inference 
enables smaller confidence intervals and more powerful p-values. We evaluate active inference on datasets from public opinion research, census analysis, and proteomics.
\end{abstract}

\section{Introduction}

In the realm of data-driven research, collecting high-quality labeled data is a continuing impediment. The impediment is particularly acute when operating under stringent labeling budgets, where the cost and effort of obtaining each label can be substantial. Recognizing these limitations, many have turned to machine learning as a pragmatic solution, leveraging it to predict unobserved labels across various fields. In remote sensing, machine learning assists in annotating and interpreting satellite imagery \cite{jean2016combining,xie2016transfer,rolf2021generalizable}; in proteomics, tools like AlphaFold \cite{jumper2021highly} are revolutionizing our understanding of protein structures; even in the realm of elections---including most major US elections---technologies combining scanners and predictive models are used as efficient tools for vote counting \cite{uselections}.
These applications reflect a growing reliance on machine learning for extracting information and knowledge from unlabeled datasets.

However, this reliance on machine learning is not without its pitfalls. The core issue lies in the inherent biases of these models. No matter how sophisticated, 
predictions lead to dubious conclusions; as such, predictions cannot fully substitute for traditional data sources such as gold-standard experimental measurements, high-quality surveys, and expert annotations. This begs the question: is there a way to effectively leverage the predictive power of machine learning while still ensuring the integrity of our inferences?

Drawing inspiration from the concept of active learning, we propose \emph{active inference}---a novel methodology for statistical inference that harnesses machine learning not as a replacement for data collection but as a strategic guide to it. The methodology uses a machine learning model to identify which data points would be most beneficial to label, thus effectively utilizing the labeling budget.
It operates on a simple yet powerful intuition: prioritize the collection of labels for data points where the model exhibits uncertainty, and rely on the model's predictions where it is confident.
Active inference constructs provably valid confidence intervals and hypothesis tests for any black-box machine learning model and any data distribution. The key takeaway is that it achieves the same level of accuracy with far fewer samples than existing baselines relying on non-adaptively-collected data. Put differently, this means that for the same number of collected  samples, active inference 
enables smaller confidence intervals and more powerful p-values. We will show in our experiments that active inference can save over $80\%$ of the sample budget required by classical inference methods (see Figure \ref{fig:budgetsave}).

Although quite different in scope, our work is inspired by the recent framework of \emph{prediction-powered inference} (PPI) \cite{angelopoulos2023prediction}. PPI assumes access to a small labeled dataset and a large unlabeled dataset, drawn i.i.d.~from the population of interest. It then asks how one can use machine learning and the unlabeled dataset to sharpen inference about population parameters depending on the distribution of labels. Our objective in this paper is different since the core of our contribution is (1) designing \emph{strategic} data collection approaches that enable more powerful inferences than collecting labels in an i.i.d.~manner, and (2) showing how to perform inference with such strategically collected data. That said, we will see that 
prediction-powered inference can be seen as a special case of our methodology: while PPI ignores the issue of strategic data collection and instead uses a trivial, uniform data collection strategy, it leverages machine learning to enhance inference in a similar way to our method. We provide a further discussion of prior work in Section~\ref{sec:related_work}.

\section{Problem description}

We introduce the formal problem setting. We observe unlabeled instances $X_1, \dots, X_n$, drawn i.i.d.~from a distribution $\P_X$. The labels $Y_i$ are unobserved, and we shall use 
$(X,Y)\sim\P = \P_X \times \P_{Y|X}$ to denote a generic feature--label pair drawn from the underlying data distribution.
We are interested in performing inference---conducting a hypothesis test or forming a confidence interval---for a parameter $\theta^*$ that depends on the distribution of the unobserved labels; that is, the parameter is a functional of $\P_X \times \P_{Y|X}$. For example, we might be interested in forming a confidence interval for the mean label, $\theta^* = \E[Y_i]$, where $Y_i$ is the label corresponding to $X_i$. Although we will primarily focus on forming confidence intervals, the standard duality between confidence intervals and hypothesis tests makes our results directly applicable to testing as well. 

We have no collected labels a priori. Rather, the goal is to efficiently and strategically acquire labels for certain points $X_i$, so that inference is as powerful as possible for a given collection budget---more so than if labels were collected uniformly at random---while also remaining valid.
We denote by $\nlab$ the number of collected labels. We assume that we are constrained to collect, on average, $\E[\nlab]\leq n_b$ labels, for some budget $n_b$. (For simplicity we bound $\E[\nlab]$, however the budget can be met with high probability, since $\nlab$ concentrates around $n_b$ at a fast rate.) Typically, $n_b\ll n$.

To guide the choice of which instances to label, we will make use of a predictive model $f$. Typically this will be a black-box machine learning model, but it could also be a hand-designed decision rule based on expert knowledge. This is the key component that will enable us to get a significant boost in power. We do not assume any knowledge of the predictive performance of $f$, or any parametric form for it. Our key takeaway is that, if we have a reasonably good model for predicting the labels $Y_i$ based on $X_i$, then we can achieve a significant boost in power compared to labeling a uniformly-at-random chosen set of instances.

We will consider two settings, depending on whether or not we update the predictive model $f$ as we gather more labels. 
\begin{itemize}
    \item The first is a \emph{batch} setting, where we simultaneously make decisions of whether or not to collect the corresponding label for all unlabeled points at once. In this setting, the model $f$ is pre-trained and remains fixed during the label collection. The batch setting is simpler and arguably more practical if we already have a good off-the-shelf predictor. 
    \item The second setting is \emph{sequential}: we go through the unlabeled points one by one and update the predictive model as we collect more data. The benefit of the second approach is that it is applicable even when we do not have access to a pre-trained model, but we have to train a model from scratch.
\end{itemize}

Our proposed active inference strategy will be applicable to all convex \emph{M-estimation} problems. This means that it handles all targets of inference $\theta^*$ that can be written as:
$$\theta^* = \argmin_{\theta} \E[\ell_\theta(X,Y)], \text{ where } (X,Y)\sim \P,$$
for a loss function $\ell_{\theta}$ that is convex in $\theta$. We denote $L(\theta)=\E[\ell_\theta(X,Y)]$ for brevity.
M-estimation captures many relevant targets, such as the mean label, quantiles of the label, and regression coefficients.

\begin{example}[Mean label]
\label{ex:mean_loss}
If $\ell_{\theta}(x,y) = \frac 1 2 (y-\theta)^2$, then the target is the mean label, $\theta^* = \E[Y]$. Note that this loss has no dependence on the features.
\end{example}

\begin{example}[Linear regression]
If $\ell_{\theta}(x,y) = \frac 1 2 (y-x^\top\theta)^2$, then $\theta^*$ is the vector of linear regression coefficients obtained by regressing $y$ on $x$, that is, the ``effect'' of $x$ on~$y$.
\end{example}

\begin{example}[Label quantile]
For a given $q\in(0,1)$, let $\ell_{\theta}(x,y) = q(y-\theta)\one\{y>\theta\} + (1-q)(\theta - y)\one\{y\leq\theta\}$ be the ``pinball'' loss. Then, $\theta^*$ is equal to the $q$-quantile of the label distribution: $\theta^* = \inf\{\theta: \P(Y\leq \theta)\geq q\}$. 
\end{example}

\section{Related work}
\label{sec:related_work}

Our work is most closely related to prediction-powered inference (PPI) and other recent works on inference with machine learning predictions \cite{angelopoulos2023prediction,angelopoulos2023ppipp,zrnic2023cross, motwani2023valid, gan2023prediction, miao2023assumption}. This recent literature in turn relates to classical work on inference with missing data and semiparametric statistics \cite{rubin1976inference, rubin1987multiple, rubin1996multiple, robins1994estimation, robins1995semiparametric, chernozhukov2018double}, as well as semi-supervised inference \cite{zhang2019semi, azriel2022semi, zhang2022high}. We consider the same set of inferential targets as in \cite{angelopoulos2023prediction,angelopoulos2023ppipp,zrnic2023cross}, building on classical M-estimation theory \cite{van2000asymptotic} to enable inference. While PPI assumes access to a small labeled dataset and a large unlabeled dataset, which are drawn i.i.d., our work is different in that it leverages machine learning in order to design \emph{adaptive} label collection strategies, which breaks the i.i.d.~structure between the labeled and the unlabeled data. That said, we shall however see that our active inference estimator will reduce to the prediction-powered estimator when we apply a trivial, uniform label collection strategy. We will demonstrate empirically that the adaptivity in label collection enables significant improvements in statistical power.

There is a growing literature on inference from adaptively collected data \cite{kato2020efficient, zhang2021statistical, cook2023semiparametric}, often focusing on data collected via a bandit algorithm. These papers typically focus on average treatment effect estimation. In contrast to our work, these works generally do not focus on how to set the data-collection policy as to achieve good statistical power, but their main focus is on providing valid inferences given a fixed data-collection policy. Notably, \citet{zhang2021statistical} study inference for M-estimators from bandit data. However, their estimators do not leverage machine learning, which is central to our work.

A substantial line of work studies adaptive experiment design \cite{robbins1952some, lai1985asymptotically, hu2006theory, list2011so, hahn2011adaptive, bhattacharya2012inferring, kasy2021adaptive, hadad2021confidence, chandak2023adaptive}, often with the goal of maximizing welfare during the experiment or identifying the best treatment. Most related to our motivation, a subset of these works \cite{list2011so, hahn2011adaptive, chandak2023adaptive} study adaptive design with the goal of efficiently estimating average treatment effects. While our motivation is not necessarily treatment effect estimation, we continue in a similar vein---collecting data adaptively with the goal of improved efficiency---with a focus on using modern, black-box machine learning to produce uncertainty estimates that can be turned into efficient label collection methods. Related variance-reduction ideas appear in stratified survey sampling \cite{nassiuma2001survey, sarndal2003model, kalton2020introduction, khan2015designing}. Our proposal can be seen as stratifying the population of interest based on the certainty of a black-box machine learning model.

Finally, our work draws inspiration from active learning, which is a subarea of machine learning centered around the observation that a machine learning model can enhance its predictive capabilities if it is allowed to choose the data points from which it learns. In particular, our setup is analogous to pool-based active learning \cite{settles2009active}. Sampling according to a measure of predictive uncertainty is a central idea in active learning~\cite{schohn2000less, tong2001support, balcan2006agnostic, joshi2009multi, hanneke2014theory, gal2017deep, ash2019deep, ren2021survey}. Since our goal is statistical inference, rather than training a good predictor, our sampling rules are different and adapt to the inferential question at hand.
More generally, we note that there is a large body of work studying ways to efficiently collect gold-standard labels, often under a budget constraint \cite{cheng2022many, zhang2023active, vishwakarma2022good}.

\section{Warm-up: active inference for mean estimation}
\label{sec:basic_idea}

We first focus on the special case of estimating the mean label, $\theta^* = \E[Y]$, in the batch setting. The intuition derived from this example carries over to all other problems.

Recall the setup: we observe $n$ i.i.d. unlabeled instances $X_1, \dots, X_n$, and we can collect labels for at most $n_b$ of them (on average).
Consider first a ``classical'' solution, which does not leverage machine learning. Given a budget $n_b$, we can simply label any arbitrarily chosen $n_b$ points. Since the instances are i.i.d., without loss of generality we can choose to label instances $\{1,\dots,n_b\}$ and compute:
\[\hat\theta^{\texttt{noML}} = \frac{1}{n_b} \sum_{i=1}^{n_b} Y_i.\]
The estimator $\hat\theta^{\texttt{noML}}$ is clearly unbiased. It is an average over $n_b$ terms, and thus its variance is equal to
\[\Var(\hat\theta^{\texttt{noML}}) =\frac{1}{n_b} \Var(Y).\]

Now, suppose that we are given a machine learning model $f(X)$, which predicts the label $Y \in \mathbb{R}$ from observed covariates $X \in \mathcal{X}$.\footnote{$\mathcal{X}$ is the set of values the covariates can take on, e.g.~$\mathbb{R}^d$.} The idea behind our active inference strategy is to increase the effective sample size by using the model's predictions on points $X$ where the model is confident and focusing the labeling budget on the points $X$ where the model is uncertain.
To implement this idea, we design a \emph{sampling rule} $\pi:\X\to[0,1]$ and collect label $Y_i$ with probability $\pi(X_i)$. The sampling rule is derived from $f$, by appropriately measuring its uncertainty. The hope is that $\pi(x) \approx 1$ signals that the model $f$ is very uncertain about instance $x$, whereas $\pi(x)\approx 0$ indicates that the model $f$ should be very certain about instance $x$. Let $\xi_i\sim\mathrm{Bern}(\pi(X_i))$ denote the indicator of whether we collect the label for point~$i$. By definition, $\nlab = \sum_{i=1}^n \xi_i$. The rule $\pi$ will be carefully rescaled to meet the budget constraint: $\E[\nlab] = \E[\pi(X)]\cdot n \leq n_b$.

Our \emph{active estimator} of the mean $\theta^*$ is given by:
\begin{align}
    \label{eqn:active-mean}
\hat\theta^{\pi} = \frac 1 n \sum_{i=1}^n \left(f(X_i) + (Y_i - f(X_i)) \frac{\xi_i}{\pi(X_i)}\right).
\end{align}
This is essentially the augmented inverse propensity weighting (AIPW) estimator \cite{robins1994estimation}, with a particular choice of propensities $\pi(X_i)$ based on the certainty of the machine learning model that predicts the missing labels. When the sampling rule is uniform, i.e.~$\pi(x) = {\nb}/{n}$ for all $x$, $\hat\theta^{\pi}$ is equal to the prediction-powered mean estimator \cite{angelopoulos2023prediction}.

It is not hard to see that $\hat\theta^{\pi}$ is unbiased: $\E[\hat\theta^{\pi}] = \theta^*$. A short calculation shows that its variance equals
\begin{align}
\label{eqn:estimator_var}
\Var(\hat\theta^{\pi}) = \frac{1}{n}\left(\Var(Y) + \E\left[(Y-f(X))^2\left(\frac{1}{\pi(X)}-1\right)\right]\right). 
\end{align}
If the model is highly accurate for all $x$, i.e. $f(X) \approx Y$, then $\Var(\hat\theta^{\pi})\approx \frac 1 n \Var(Y)$, which is far smaller than $\Var(\hat\theta^{\texttt{noML}})$ since $n_b\ll n$. Of course, $f$ will never be perfect and accurate for all instances $x$. For this reason, we will aim to choose $\pi$ such that $\pi$ is small when $f(X)\approx Y$ and large otherwise, so that the relevant term $(Y-f(X))^2\left({\pi}^{-1}(X) -1\right)$ is always small (of course, subject to the sampling budget constraint). For example, for instances for which the predictor is correct, i.e. $f(X) = Y$, we would ideally like to set $\pi(X) = 0$ as this incurs no additional variance.



We note that the variance reduction of active inference compared to the ``classical'' solution also implies that the resulting confidence intervals get smaller. This follows because interval width scales with the standard deviation for most standard intervals (e.g., those derived from the central limit theorem).


Finally, we explain how to set the sampling rule $\pi$. The rule will be derived from a measure of \emph{model uncertainty} $u(x)$ and we shall provide different choices of $u(x)$ in the following paragraphs. At a high level, one can think of $u(X_i)$ as the model's best guess of $|Y_i - f(X_i)|$. We will choose $\pi(x)$ proportional to $u(x)$, that is, $\pi(x)\propto u(x)$, normalized to meet the budget constraint. Intuitively, this means that we want to focus our data collection budget on parts of the covariate space where the model is expected to make the largest errors. Roughly speaking, we will set $\pi(x) = \frac{u(x)}{\E[u(X)]}\cdot \frac{\nb}{n}$; this implies $\E[\nlab] = \E[\pi(X)]\cdot n \leq \nb$. (This is an idealized form of $\pi(x)$ because $\E[u(X)]$ cannot be known exactly, though it can be estimated very accurately from the unlabeled data; we will formalize this in the following section.)

We will take two different approaches for choosing the uncertainty $u(x)$, depending on whether we are in a regression or a classification setting.

\paragraph{Regression uncertainty} In regression, we explicitly train a model $u(x)$ to predict $|f(X_i) - Y_i|$ from $X_i$. We note that we aim to predict only the magnitude of the error and not the directionality. In the batch setting, we typically have historical data of $(X,Y)$ pairs that are used to train the model $f$. We thus
train $u(x)$ on this historical data, by setting $|f(X) - Y|$ as the target label for instance $X$. The data used to train $u$ should ideally be disjoint from the data used to train $f$ to avoid overoptimistic estimates of uncertainty. We will typically use data splitting to avoid this issue, though there are more data efficient solutions such as cross-fitting. Notice that access to historical data will only be important in the batch setting, as assumed in this section. In the sequential setting we will be able to train $u(x)$ gradually on the collected data.

\paragraph{Classification uncertainty} 
Next we look at classification, where $Y$ is supported on a discrete set of values. Our main focus will be on binary classification, where $Y\in\{0,1\}$. In such cases, our target is $\theta^* = \P(Y=1)$. We might care about $\E[Y]$ more generally when $Y$ takes on $K$ distinct values (e.g., $K$ distinct ratings in a survey, $K$ distinct qualification levels, etc).

In classification, $f(x)$ is usually obtained as the ``most likely'' class. If $K$ is the number of classes, we have $f(x) = \argmax_{i\in[K]}  p_i(x)$, for some probabilistic output ${p}(x) = ( p_1(x),\dots, p_K(x))$ which satisfies $\sum_{i=1}^K  p_i(x) = 1$. For example, ${p}(x)$ could be the softmax output of a neural network given input $x$. We will measure the uncertainty as:
\begin{equation}
\label{eqn:classification_uncertainty}
u(x) = \frac{K}{K-1} \cdot \left(1 - \max_{i\in[K]}  p_i(x)\right).
\end{equation}
In binary classification, this reduces to $u(x) = 2\min\{p(x), 1-p(x)\}$, where we use $p(x)$ to denote the raw classifier output in $[0,1]$. Therefore, $u(x)$ is large when $ p(x)$ is close to uniform, i.e. $\max_i  p_i(x)\approx 1/K$. On the other hand, if the model is confident, i.e. $\max_i  p_i(x)\approx 1$, the uncertainty is close to zero.

\section{Batch active inference}
\label{sec:batch}

Building on the discussion from Section \ref{sec:basic_idea}, we provide formal results for active inference in the batch setting. Recall that in the batch setting we observe i.i.d. unlabeled points $X_1,\dots,X_n$, all at once. We consider a family of sampling rules $\pi_\eta(x) = \eta \, u(x)$,  where $u(x)$ is the chosen uncertainty measure and $\eta \in \cH\subseteq \R^+$ is a tuning parameter. We will discuss ways of choosing $u(x)$ in Section \ref{sec:optimal_rule}. The role of the tuning parameter is to scale the sampling rule to the sampling budget.
We choose
\begin{equation}
\label{eqn:eta-hat}
\hat\eta = \max\left\{\eta \in \mathcal H : \eta\, \sum_{i=1}^n  u(X_i) \leq n_b \right\},
\end{equation}
and deploy $\pi_{\hat\eta}$ as the sampling rule.
With this choice, we have 
$$\E[\nlab] = \E\left[\sum_{i=1}^n \hat\eta \,  u(X_i) \right]\leq n_b;$$ 
therefore, $\pi_{\hat\eta}$ meets the label collection budget. We denote $\hat\theta^\eta \equiv \hat\theta^{\pi_\eta}$.


\paragraph{Mean estimation}
We first explain how to perform inference for mean estimation in Proposition \ref{prop:mean_batch_inference}. Recall the active mean estimator: 
\begin{equation}
\label{eq:mean_est_batch}
\hat\theta^{\hat\eta} = \frac 1 n \sum_{i=1}^n \left(f(X_i) + (Y_i - f(X_i))\frac{\xi_i}{\pi_{\hat\eta}(X_i)} \right),
\end{equation}
where $\xi_i\sim\mathrm{Bern}(\pi_{\hat\eta}(X_i))$. Following standard notation, $z_q$ below denotes the $q$th quantile of the standard normal distribution.

\begin{prop}
\label{prop:mean_batch_inference}
Suppose that there exists $\eta^*\in\cH$ such that $\P(\hat\eta \neq \eta^*)\to 0$. Then
$$\sqrt{n}(\hat\theta^{{\hat\eta}} - \theta^*)\cd \cN(0,\sigma_*^2),$$
    where $\sigma_*^2 = \Var(f(X) + (Y-f(X))\frac{\xi^{\eta^*}}{\pi_{\eta^*}(X)})$ and $\xi^{\eta^*}\sim\mathrm{Bern}(\pi_{\eta^*}(X))$. Consequently, for any $\hat\sigma^2\cp\sigma_*^2$, $\C_{\alpha} = (\hat\theta^{{\hat\eta}} \pm z_{1-\alpha/2}\frac{\hat\sigma}{\sqrt{n}})$ is a valid $(1-\alpha)$-confidence interval: $$\lim_{n \rightarrow \infty}  \,\, \P(\theta^*\in\C_{\alpha}) = 1-\alpha.$$
\end{prop}

A few remarks about Proposition \ref{prop:mean_batch_inference} are in order:
first, the consistency condition $\P(\hat\eta\neq\eta^*)\to 0$ is easily ensured if ${n_b}/{n}$ has a limit $p\in(0,1)$, that is, if $n_b$ is asymptotically proportional to $n$. Then, as long as the space of tuning parameters $\cH$ is discrete and there is no $\eta\in\cH$ such that $\eta \, \E[ u(X)] = p$ exactly, the consistency condition is met. Second, obtaining a consistent variance estimate $\hat\sigma^2$ is straightforward, as one can simply take the empirical variance of the increments in the estimator \eqref{eq:mean_est_batch}.

We note that, while our main results will all focus on asymptotic confidence intervals, some of our results have direct non-asymptotic and time-uniform analogues; see Section \ref{sec:nonasymptotic}.

\paragraph{General M-estimation}
Next, we turn to general convex M-estimation. Recall  this means that we can write $\theta^* = \argmin_\theta L(\theta) = \argmin_\theta \E[\ell_\theta(X,Y)]$, for a convex loss $\ell_\theta$. To simplify notation, let $\ell_{\theta,i} = \ell_\theta(X_i,Y_i)$, $\ell_{\theta,i}^f = \ell_\theta(X_i,f(X_i))$. We similarly use $\nabla \ell_{\theta,i}$ and $\nabla \ell_{\theta,i}^f$. 
 For a general sampling rule $\pi$, our \emph{active estimator} is defined as
\begin{align}
\label{eqn:general_estimator}
&\hat \theta^\pi = \argmin_\theta L^\pi(\theta), \text{ where } L^\pi(\theta) = \frac 1 n \sum_{i=1}^n \left( \ell_{\theta,i}^f + (\ell_{\theta,i} - \ell_{\theta,i}^f)\frac{\xi_i}{\pi(X_i)}\right).
\end{align}
As before, $\xi_i\sim\mathrm{Bern}(\pi(X_i))$. When $\pi$ is the uniform rule, $\pi(x) = {\nb}/{n}$, the estimator \eqref{eqn:general_estimator} equals the general prediction-powered estimator from \cite{angelopoulos2023ppipp}. Notice that the loss estimate $L^\pi(\theta)$ is unbiased: $\E[L^\pi(\theta)] = L(\theta)$.
We again scale the sampling rule $\pi_\eta(x) = \eta\, u(x)$ according to the sampling budget, as in Eq.~\eqref{eqn:eta-hat}.

We next show asymptotic normality of $\hat\theta^{{\hat\eta}}$ for general targets $\theta^*$ which, in turn, enables inference. The result essentially follows from the usual asymptotic normality for M-estimators \citep[Ch.~5]{van2000asymptotic}, with some necessary modifications to account for the data-driven selection of $\hat\eta$. We require standard, mild smoothness assumptions on the loss $\ell_\theta$, formally stated in Ass.~\ref{ass:smooth_loss} in the Appendix.

\begin{theorem}[CLT for batch active inference]
\label{thm:m_est_batch}
Assume the loss is smooth (Ass.~\ref{ass:smooth_loss}) and define the Hessian $H_{\theta^*} = \nabla^2 \E[\ell_{\theta^*}(X,Y)]$. Suppose that there exists $\eta^*\in\cH$ such that $\P(\hat\eta\neq \eta^*)\to 0$. Then, if $\hat\theta^{\eta^*}\cp \theta^*$, we have 
$$\sqrt{n}(\hat\theta^{{\hat\eta}} - \theta^*)\cd \cN(0,\Sigma_*), \text{ where }$$
   $\Sigma_* = H_{\theta^*}^{-1} \Var\left(\nabla \ell_{\theta^*,i}^f  + \left(\nabla \ell_{\theta^*,i} - \nabla \ell_{\theta^*,i}^f \right) \frac{\xi^{\eta^*}}{\pi_{\eta^*}(X)} \right)H_{\theta^*}^{-1}$. Consequently, for any $\hat\Sigma\cp \Sigma_*$, $\C_\alpha = (\hat\theta^{{\hat\eta}}_j \pm z_{1-\alpha/2}\sqrt{\frac{\hat\Sigma_{jj}}{n}})$ is a valid $(1-\alpha)$-confidence interval for $\theta^*_j$:
   $$\lim_{n \rightarrow \infty} \,\, \P(\theta^*_j\in \C_\alpha) = 1-\alpha.$$
\end{theorem}

The remarks following Proposition \ref{prop:mean_batch_inference} again apply: the consistency condition on $\hat\eta$ is easily ensured if ${n_b}/{n}$ has a limit, and $\hat\Sigma$ admits a simple plug-in estimate by replacing all quantities with their empirical counterparts. The consistency condition on $\hat\theta^{\eta^*}$ is a standard requirement for analyzing M-estimators \citep[see][Ch.~5]{van2000asymptotic}; it is studied and justified at length in the literature and we shall therefore not discuss it in close detail. We however remark that it can be deduced if the empirical loss $L^\pi(\theta)$ is almost surely convex or if the parameter space is compact. The empirical loss $L^\pi(\theta)$ is convex in a number of cases of interest, including means and generalized linear models; for the proof, see \cite{angelopoulos2023ppipp}.

\section{Sequential active inference}
\label{sec:sequential}

In the batch setting we observe all data points $X_1,\dots,X_n$ at once and fix a predictive model $f$ and sampling rule $\pi$ that guide our choice of which labels to collect. An arguably more natural data collection strategy would operate in an online manner: as we collect more labels, we iteratively update the model and our strategy for which labels to collect next. This allows for further efficiency gains over using a fixed model throughout, as the latter ignores knowledge acquired during the data collection. For example, if we are conducting a survey and we collect responses from members of a certain demographic group, it is only natural that we update our sampling rule to reflect the fact that we have more knowledge and certainty about that demographic group.

Formally, instead of having a fixed model $f$ and rule~$\pi$, we go through our data sequentially. At step $t\in~\{1,\dots,n\}$, we observe data point $X_t$ and collect its label with probability $\pi_t(X_t)$, where $\pi_t(\cdot)$ is based on the uncertainty of model $f_t$. The model $f_t$ can be fine-tuned on all information observed up to time $t$; formally, we require that $f_t,\pi_t\in\F_{t-1}$, where $\F_{t}$ is the $\sigma$-algebra generated by the first $t$ points $X_s$, $1 \le s \le t$,  their labeling decisions $\xi_s$, and their labels $Y_s$, if observed:
$\F_{t}=\sigma((X_1, Y_1\xi_1,\xi_1),\dots,(X_{t},Y_{t}\xi_{t}, \xi_{t})).$
(Note that $Y_t\xi_t = Y_t$ if and only if $\xi_t=1$; otherwise, $Y_t\xi_t = \xi_t = 0$.)
We will again calibrate our decisions of whether to collect a label according to a budget on the sample size $\nb$. We denote by $\nlabt$ the number of labels collected up to time $t$.

Inference in the sequential setting is more challenging than batch inference because the data points $(X_t,Y_t,\xi_t), t\in [n]$, are dependent; indeed, the purpose of the sequential setting is to leverage previous observations when deciding on future labeling decisions. We will construct estimators that respect a \emph{martingale} structure, which will enable tractable inference 
via the martingale central limit theorem~\cite{dvoretzky1972asymptotic}. This resembles the approach taken by \citet{zhang2021statistical} (though our estimators are quite different due to the use of machine learning predictions).

\paragraph{Mean estimation}
We begin by focusing on the mean. If we take $\ell_\theta$ to be the squared loss as in Example~\ref{ex:mean_loss}, we obtain the sequential active mean estimator:
$$\hat\theta^{\vv{\pi}} = \frac 1 n \sum_{t=1}^n \Delta_t, \qquad \Delta_t = f_t(X_t) + (Y_t - f_t(X_t)) \frac{\xi_t}{\pi_t(X_t)}.$$
We note that $\Delta_t$ are \emph{martingale increments}; they share a common conditional mean $\E[\Delta_t|\F_{t-1}] = \theta^*$, and they are $\F_t$-measurable, $\Delta_t \in \F_t$. We let $\sigma^2_t =  V(f_t,\pi_t) = \Var(\Delta_t | f_t, \pi_t)$ denote the conditional variance of the increments.

To show asymptotic normality of $\hat\theta^{\vv{\pi}}$, we shall require the Lindeberg condition, whose statement we defer to the Appendix. It is a standard assumption for proving central limit theorems when the increments are not i.i.d.. Roughly speaking, the Lindeberg condition requires that the increments do not have very heavy tails; it prevents any increment from having a disproportionately large contribution to the overall variance. 

\begin{prop}
\label{prop:sequential_mean}
Suppose $\frac 1 n \sum_{t=1}^n \sigma^2_t \stackrel{p}{\to} \sigma^2_* = V(f_*, \pi_*)$, for some fixed model--rule pair $(f_*, \pi_*)$, and that the increments $\Delta_t$ satisfy the Lindeberg condition (Ass.~\ref{ass:lindeberg_mean}). Then
$$\sqrt{n}(\hat\theta^{\vv{\pi}} - \theta^*) \cd \mathcal{N}(0,  \sigma^2_*).$$
Consequently, for any $\hat\sigma^2\cp\sigma_*^2$, $\C_\alpha = (\hat\theta^{\vv{\pi}} \pm z_{1-\alpha/2}\frac{\hat\sigma}{\sqrt{n}})$ is a valid $(1-\alpha)$-confidence interval: $$\lim_{n \rightarrow \infty} \,\,  \P(\theta^* \in \C_\alpha) = 1-\alpha.$$
\end{prop}

Intuitively, Proposition \ref{prop:sequential_mean} requires that the model $f_t$ and sampling rule $\pi_t$ converge. For example, a sufficient condition for $\frac 1 n \sum_{t=1}^n \sigma^2_t \stackrel{p}{\to} \sigma^2_*$ is $V(f_n,\pi_n) \stackrel{L^1}{\to} V(f_*,\pi_*)$. Since the sampling rule is typically based on the model, it makes sense that it would converge if $f_t$ converges. At the same time, it makes sense for $f_t$ to gradually stop updating after sufficient accuracy is achieved.

\paragraph{General M-estimation}
We generalize Proposition~\ref{prop:sequential_mean} to all convex M-estimation problems.
The general version of our sequential active estimator takes the form
\begin{align}
\label{eqn:sequential_est_general}
&\hat\theta^{\vv{\pi}} = \argmin_\theta L^{\vv{\pi}}(\theta), \text{ where } L^{\vv{\pi}}(\theta)= \frac 1 n \sum_{t=1}^n L_t(\theta), \qquad  L_t(\theta) = \ell_{\theta,t}^{f_t} + (\ell_{\theta,t} - \ell_{\theta,t}^{f_t}) \frac{\xi_t}{\pi_t(X_t)}.
\end{align}
Let $V_{\theta,t} = V_{\theta}(f_t,\pi_t) = \Var\left( \nabla L_t(\theta)| f_t, \pi_t\right)$.
We will again require that $(f_t,\pi_t)$ converge in an appropriate sense. 

\begin{theorem}[CLT for sequential active inference]
\label{thm:sequential_general}
Assume the loss is smooth (Ass.~\ref{ass:smooth_loss}) and define the Hessian $H_{\theta^*} = \nabla^2 \E[\ell_{\theta^*}(X,Y)]$. Suppose also that $\frac 1 n \sum_{t=1}^n V_{\theta^*,t} \stackrel{p}{\to} V_{*} = V_{\theta^*}(f_*,\pi_*)$ entry-wise for some fixed model--rule pair $(f_*,\pi_*)$, that the increments $L_t(\theta)$ satisfy the Lindeberg condition (Ass.~\ref{ass:lindeberg_general}), and that the Hessian is locally Lipschitz in a neighborhood of $\theta^*$. Then, if $\hat\theta^{\vv{\pi}}\cp \theta^*$, we have 
$$\sqrt{n}(\hat\theta^{\vv{\pi}} - \theta^*)\cd \cN(0,\Sigma_*),$$
    where $\Sigma_* = H_{\theta^*}^{-1} V_* H_{\theta^*}^{-1}$. Consequently, for any $\hat\Sigma\cp \Sigma_*$, $\C_\alpha = (\hat\theta^{\vv{\pi}}_j \pm z_{1-\alpha/2}\sqrt{\frac{\hat\Sigma_{jj}}{n}})$ is a valid $(1-\alpha)$-confidence interval for $\theta^*_j$: 
    $$\lim_{n \rightarrow \infty} \,\,  \P(\theta^*_j\in \C_\alpha) = 1-\alpha.$$
\end{theorem}

The conditions of Theorem \ref{thm:sequential_general} are largely the same as in Theorem \ref{thm:m_est_batch}; the main difference is the requirement of convergence of the model--sampling rule pairs, which is similar to the analogous condition of Proposition~\ref{prop:sequential_mean}. 

Proposition \ref{prop:sequential_mean} and Theorem \ref{thm:sequential_general} apply to any sampling rule $\pi_t$, as long as the variance convergence requirement is met. We discuss ways to set $\pi_t$ so that the sampling budget $\nb$ is met. Our default will be to ``spread out'' the budget $\nb$ over the $n$ observations. We will do so by having an ``imaginary'' budget for the expected number of collected labels by step $t$, equal to $n_{b,t} = {t} \nb/n$. Let $n_{\Delta,t} = n_{b,t} - \nlabtminone $ denote the remaining budget at step $t$. We derive a measure of uncertainty $u_t$ from model $f_t$, as before, and let 
\begin{equation}
\label{eqn:sequential_rule}
\pi_{t}(x) = \min\left\{\eta_t \, u_{t}(x), n_{\Delta,t} \right\}_{[0,1]},
\end{equation}
where $\eta_t$ normalizes $u_t(x)$ and the subscript $[0,1]$ denotes clipping to $[0,1]$. The normalizing constant $\eta_t$ can be arbitrary, but we find it helpful to set it roughly as $\eta_t = {n_b}/({n \, \E[u_t(X)]})$ and this is what we do in our experiments, with the proviso that we substitute $\E[u_t(X)]$ with its empirical approximation.
In words, the sampling probability is high if the uncertainty is high \emph{and} we have not used up too much of the sampling budget thus far.
Of course, if the model consistently estimates low uncertainty $u_t(x)$ throughout, the budget will be underutilized. For this reason, to make sure we use up the budget in practice, we occasionally set $\pi_t(x) = (n_{\Delta,t})_{[0,1]}$ regardless of the reported uncertainty. This periodic deviation from the rule~\eqref{eqn:sequential_rule} is consistent with the variance convergence conditions required for Proposition~\ref{prop:sequential_mean} and Theorem~\ref{thm:sequential_general} to hold.

\section{Choosing the sampling rule}
\label{sec:optimal_rule}

We have seen how to perform inference given an abstract sampling rule, and argued that, intuitively, the sampling rule should be calibrated to the uncertainty of the model's predictions. Here we argue that this is in fact the \emph{optimal} strategy.
In particular, we derive an ``oracle'' rule, which optimally sets the sampling probabilities so that the variance of $\hat\theta^\pi$ is minimized. While the oracle rule cannot be implemented since it depends on unobserved information, it provides an ideal that our algorithms will try to approximate. We discuss ways of tuning the approximations to make them practical and powerful.

\subsection{Oracle sampling rules}

We begin with the optimal sampling rule for mean estimation.  We then state the optimal rule for general M-estimation and instantiate it for generalized linear models.

\paragraph{Mean estimation} Recall the expression for $\Var(\hat\theta^\pi)$~\eqref{eqn:estimator_var}. Given that $\E\left[{\pi^{-1}(X)} (Y-f(X))^2\right]$ is the only term that depends on $\pi$,
we define the oracle rule as the solution to:
\begin{align}
\label{eqn:oracle_utility}	
\underset{\pi}{\min} \quad \E\left[\frac{1}{\pi(X)}(Y-f(X))^2\right] \text{ s.t. } \E[\pi(X)]\leq \frac{n_b}{n}.
\end{align} 
The optimization problem \eqref{eqn:oracle_utility} appears in a number of other topics, including importance sampling \citep[Ch.~9]{owen_mcbook}, constrained utility optimization \cite{balcan2014learning}, and, relatedly to our work, survey sampling \cite{sarndal1980pi}. The optimality conditions of \eqref{eqn:oracle_utility} show that its solution $\piopt$ satisfies:
\begin{equation*}
\piopt(X) \propto \sqrt{\E[(Y - f(X))^2|X]},
\end{equation*}
where $\propto$ ignores the normalizing constant required to make $\E[\piopt(X)]\leq {n_b}/{n}$. Therefore, the optimal sampling rule is one that samples data points according to the expected magnitude of the model error: the larger the model error, the higher the probability of sampling should be.
Of course, $\E[(Y-f(X))^2|X]$ cannot be known since the label distribution is unknown, and that is why we call $\piopt$ an oracle.

To develop intuition, it is instructive to consider an even more powerful oracle $\tilde \pi_{\mathrm{opt}}(X,Y)$ that is allowed to depend on $Y$. To be clear, we would commit to the same functional form as in \eqref{eqn:active-mean} and would seek to 
minimize $\Var(\hat\theta^\pi)$ while allowing the sampling probabilities to depend on both $X$ \emph{and} $Y$. In this case, by the same argument we conclude that
\begin{equation}
\label{eqn:optrule_mean}\tilde \pi_{\mathrm{opt}}(X,Y) \propto |Y-f(X)|.
\end{equation}
The perspective of allowing the oracle to depend on both $X$ and $Y$ is directly prescriptive: a natural way to approximate the rule $\tilde \pi_{\mathrm{opt}}$ is to train an arbitrary black-box model $u$ on historical $(X,Y)$ pairs to predict $|Y-f(X)|$ from $X$. We provide further practical guidelines for sampling rules at the end of this section.

\paragraph{General M-estimation} In the case of general M-estimation, we cannot hope to minimize the variance of $\hat\theta^\pi$ at a fixed sample size $n$ since the finite-sample distribution of $\hat\theta^\pi$ is not tractable. However, we can find a sampling rule that minimizes the \emph{asymptotic} variance of $\hat\theta^\pi$. Since the estimator is potentially multi-dimensional, to make the problem well-posed we assume that we want to minimize the asymptotic variance of a single coordinate $\hat\theta^{\pi}_j$ (for example, one coefficient in a multi-dimensional regression). Recall the expression for the asympotic covariance $\Sigma_*$ from Theorem~\ref{thm:m_est_batch}.
A short derivation shows that
$$\Sigma_{*,jj} =  \E\left[ \left(\left(\nabla \ell_{\theta^*,i} - \nabla \ell_{\theta^*,i}^f \right)^\top h^{(j)}\right)^2  \cdot \frac{1}{\pi(X)}\right] + C,$$
where $h^{(j)}$ is the $j$-th column of $H_{\theta^*}^{-1}$ and $C$ is a term that has no dependence on the sampling rule $\pi$. Therefore, by the same theory as for mean estimation, the ideal rule $\piopt(X)$ would be
$$\piopt(X) \propto \sqrt{\E[(\left(\nabla \ell_{\theta^*}(X,Y) - \nabla \ell_{\theta^*}(X,f(X)) \right)^\top  h^{(j)})^2|X]}.$$
This recovers $\piopt$ for the mean, since $\nabla \ell_{\theta^*}(x,y) = \theta^* - y$ and $h^{(j)} = 1$ for the squared loss. Our measure of uncertainty $u(x)$ should therefore approximate the errors of the predicted gradients along the $h^{(j)}$ direction.

\paragraph{Generalized linear models (GLMs)}
We simplify the general solution $\piopt$ in the case of generalized linear models (GLMs). We define GLMs as M-estimators whose loss function takes the form
$$\ell_{\theta}(x,y) = -\log p_\theta(y|x) = -yx^\top\theta + \psi(x^\top\theta),$$
for some convex log-partition function $\psi$. This definition recovers linear regression by taking $\psi(s) = \frac 1 2 s^2$ and logistic regression by taking $\psi(s) = \log(1+e^s)$. By the definition of the GLM loss, we have $\nabla \ell_{\theta^*}(x,y) - \nabla \ell_{\theta^*}(x,f(x)) = (f(x) - y) x$ and, therefore,
\begin{equation*}
\piopt(X) \propto \sqrt{\E[(f(X) - Y)^2|X]} \cdot |X^\top  h^{(j)}|,
\end{equation*}
where the Hessian is equal to $H_{\theta^*} = \E[\psi''(X^\top\theta_*)XX^\top]$ and $h^{(j)}$ is the $j$-th column of $H_{\theta^*}^{-1}$. In linear regression, for instance, $H_{\theta^*} = \E[XX^\top]$. Again, we see that the model errors play a role in determining the optimal sampling. In particular, again considering the more powerful oracle $\tilde \pi_{\mathrm{opt}}(X,Y)$ that is allowed to set the sampling probabilities according to both $X$ and $Y$, we get
\begin{equation}
\label{eqn:optimal_glm}\tilde \pi_{\mathrm{opt}}(X,Y) \propto |f(X) - Y| \cdot |X^\top h^{(j)}|.
\end{equation}
Therefore, as in the case of the mean,
our measure of uncertainty will aim to predict $|f(X) - Y|$ from $X$ and plug those predictions into the above rule.

\subsection{Practical sampling rules}
\label{sec:practical_rules}

As explained in Section \ref{sec:batch} and Section \ref{sec:sequential}, our sampling rule $\pi(x)$ will be derived from a measure of uncertainty $u(x)$. As clear from the preceding discussion, the right notion of uncertainty should measure a notion of error dependent on the estimation problem at hand. In particular, we hope to have $u(X)\approx | \left(\nabla \ell_{\theta^*}(X,Y) - \nabla \ell_{\theta^*}(X,f(X)) \right)^\top  h^{(j)}|$. For GLMs and means, in light of Eq.~\eqref{eqn:optrule_mean} and Eq.~\eqref{eqn:optimal_glm}, this often boils down to training a predictor of $|f(X) - Y|$ from $X$ and, in the case of GLMs, using a plug-in estimate of the Hessian. This is what we do in our experiments (except in the case of binary classification where we simply use the uncertainty from Eq.~\eqref{eqn:classification_uncertainty}).

Of course, the learned predictor of model errors cannot be perfect; as a result, $\pi(x)\propto u(x)$ cannot naively be treated as the oracle rule $\piopt$. For example, the model might mistakenly estimate (near-)zero uncertainty ($u(X) \approx 0$) when $|f(X) - Y|$ is large, which would blow up the estimator variance. To fix this issue, we find that it helps to stabilize the rule $\pi(x)\propto u(x)$ by mixing it with a uniform rule.

Denote the uniform rule by $\pi^{\mathrm{unif}}(x) = {n_b}/{n}$. Clearly the uniform rule meets the budget constraint, since $n\, \E[\pi^{\mathrm{unif}}(X)]  = n_b$. For a fixed $\tau \in [0,1]$ and $\pi(x)\propto u(x)$, we define the $\tau$-mixed rule as 
$$\pi^{(\tau)}(x) = (1-\tau) \cdot \pi(x) + \tau \cdot \pi^{\mathrm{unif}}(x).$$
Any positive value of $\tau$ ensures that $\pi^{(\tau)}(x) > 0$ for all $x$, avoiding instability due to small uncertainty estimates $u(x)$. When historical data is available, one can tune $\tau$ by optimizing the empirical estimate of the (asymptotic) variance of $\hat\theta^{\pi^{(\tau)}}$ given by Theorem \ref{thm:m_est_batch}. For example, in the case of mean estimation, this would correspond to solving:
\begin{equation}
\label{eqn:tau-tuning}\hat\tau = \argmin_{\tau\in[0,1]} \sum_{i=1}^{n_h} \frac{1}{\pi^{(\tau)}(X_i^h)} (Y_i^h - f(X_i^h))^2,
\end{equation}
where $(X_i^h, Y_i^h),\dots,(X_{n_n}^h, Y_{n_h}^h)$ are the historical data points.
Otherwise, one can set $\tau$ to be any user-specified constant. In our experiments, in the batch setting we tune $\tau$ on historical data when such data is available. In the sequential setting we simply set $\tau=0.5$ as the default.

\section{Experiments}

We evaluate active inference on several problems and compare it to two baselines.

The first baseline replaces active sampling with the uniformly random sampling rule $\pi^{\mathrm{unif}}$.  
Importantly, this baseline still uses machine learning predictions $f(X_i)$ and 
corresponds to prediction-powered inference (PPI) \cite{angelopoulos2023prediction}. Formally, the prediction-powered estimator is given by
$$\hat\theta^{\texttt{PPI}} = \argmin_\theta L^{\texttt{PPI}}(\theta), \text{ where } L^{\texttt{PPI}}(\theta) = \frac{1}{n} \sum_{i=1}^n \ell_\theta(X_i,f(X_i)) + \frac{1}{n_b} \sum_{i=1}^n \left(\ell_\theta(X_i,Y_i) - \ell_\theta(X_i,f(X_i))\right) \xi_i,$$
where $\xi_i\sim\mathrm{Bern}(\frac{\nb}{n})$. This estimator can be recovered as a special case of estimator \eqref{eqn:general_estimator}. 
The purpose 
of this comparison is to quantify the benefits of machine-learning-driven data collection. In the rest of this section we refer to this baseline as the ``uniform'' baseline because the only difference from our estimator is that it replaces active sampling with uniform sampling.

The second baseline removes machine learning altogether and computes the ``classical'' estimate based on uniformly random sampling, 
$$\hat\theta^{\texttt{noML}} = \argmin_\theta \frac{1}{\nb} \sum_{i=1}^n \ell_\theta(X_i,Y_i) \xi_i,$$
where $\xi_i\sim\mathrm{Bern}(\frac{\nb}{n})$.
This baseline serves to evaluate the cumulative benefits of machine learning for data collection \emph{and} inference combined. We refer to this baseline as the ``classical'' baseline, or classical inference, in the rest of this section.

\begin{algorithm}[t]
\caption{Batch active inference}
\label{alg:batch}
\begin{algorithmic}[1]
\Require unlabeled data $X_1,\dots,X_n$, sampling budget $n_b$, predictive model $f$, error level $\alpha\in(0,1)$
\State Choose uncertainty measure $u(x)$ based on $f$
\State Let $\pi(x) = \hat\eta~u(x)$, where $\hat\eta = \frac{n_b}{n\hat{\E}[u(X)]}$; let $\pi^{\mathrm{unif}} = \frac{n_b}{n}$
\State Select $\tau\in(0,1)$ and choose sampling rule $\pi^{(\tau)}(x) = (1-\tau)\cdot \pi(x) + \tau \cdot \pi^{\mathrm{unif}}$
\State Sample labeling decisions $\xi_i \sim \mathrm{Bern}(\pi^{(\tau)}(X_i)), i \in[n]$
\State Collect labels $\{Y_i : \xi_i = 1\}$
\State Compute batch active estimator $\hat\theta^{\pi^{(\tau)}}$ (Eq.~\eqref{eqn:general_estimator})
\end{algorithmic}
\end{algorithm}

\begin{algorithm}[t]
\caption{Sequential active inference}
\label{alg:seq}
\begin{algorithmic}[1]
\Require unlabeled data $X_1,\dots,X_n$, sampling budget $n_b$, initial predictive model $f_1$, error level $\alpha\in(0,1)$, fine-tuning batch size $B$
\State Set $\mathcal{D}^{\mathrm{tune}} \leftarrow \emptyset$
\For{$t = 1,\dots,n$}
\State Choose uncertainty measure $u_t(x)$ for $f_t$
\State Set $\pi_t(x)$ as in Eq.~\eqref{eqn:sequential_rule} with $\eta_t=\frac{n_b}{n \hat \E[u_t(X)]}$; let $\pi^{\mathrm{unif}} = \frac{n_b}{n}$
\State Select $\tau \in(0,1)$ and choose sampling rule $\pi^{(\tau)}_t(x) = (1-\tau)\cdot \pi_t(x) + \tau \cdot \pi^{\mathrm{unif}}$
\State Sample labeling decision $\xi_t \sim \mathrm{Bern}(\pi^{(\tau)}_t(X_t))$
\If{$\xi_t=1$}
\State Collect label $Y_t$
\State $\mathcal{D}^{\mathrm{tune}} \leftarrow \mathcal{D}^{\mathrm{tune}} \cup \{(X_t, Y_t)\}$
\If{$|\mathcal{D}^{\mathrm{tune}}| = B$}
\State Fine-tune model on $\mathcal{D}^{\mathrm{tune}}$: $f_{t+1} = \texttt{finetune}(f_{t}, \mathcal{D}^{\mathrm{tune}})$
\State Set $\mathcal{D}^{\mathrm{tune}} \leftarrow \emptyset$
\Else 
\State $f_{t+1} \leftarrow f_{t}$
\EndIf
\Else
\State $f_{t+1} \leftarrow f_{t}$
\EndIf
\EndFor
\State Compute sequential active estimator $\hat\theta^{\vv{\pi}^{(\tau)}}$ (Eq.~\eqref{eqn:sequential_est_general})
\end{algorithmic}
\end{algorithm}

For all methods we compute standard confidence intervals based on asymptotic normality. The target error level is $\alpha=0.1$ throughout. We report the average interval width and coverage for varying sample sizes $\nb$, averaged over $1000$ and $100$ trials for the batch and sequential settings, respectively. We plot the interval width on a log--log scale. We also report the percentage of budget saved by active inference relative to the baselines when the methods are matched to be equally accurate. More precisely, for varying $n_b$ we compute the average interval width achieved by the uniform and classical baselines; then, we look for the budget size $n_b^{\mathrm{active}}$ for which active inference achieves the same average interval width, and report $(n_b - n_b^{\mathrm{active}})/n_b \cdot 100\%$ as the percentage of budget saved.

The batch and sequential active inference methods used in our experiments are outlined in Algorithm \ref{alg:batch} and Algorithm \ref{alg:seq}, respectively. Each application will specify the general parameters from the algorithm statements.
We defer some experimental details, such as the choices of $\tau$, to Appendix \ref{app:exps}. Code for reproducing the experiments is available at \url{https://github.com/tijana-zrnic/active-inference}.

\subsection{Post-election survey research}
\label{sec:election}

We apply active inference to survey data collected by the Pew Research Center following the 2020 United States presidential election \cite{atp79}. We focus on one specific question in the survey, aimed at gauging people's approval of the presidential candidates' political messaging following the election. The target of inference is the average approval rate of Joe Biden's (Donald Trump's, respectively) political messaging. Approval is encoded as a binary response, $Y_i\in\{0,1\}$.

The respondents---a nationally representative pool of US adults---provide background information such as age, gender, education, political affiliation, etc. We show that, by training a machine learning model to predict people's approval from their background information and measuring the model's uncertainty, we can allocate the per-question budget in a way that achieves higher statistical power than uniform allocation. Careful budget allocation is important, because Pew pays each respondent proportionally to the number of questions they answer.

We use half of all available data for the analysis; for the purpose of evaluating coverage, we take the average approval on all available data as the ground truth $\theta^*$. To obtain the predictive model $f$, we train an XGBoost model \cite{chen2016xgboost} on the half of the data not used for the analysis. Since approval is encoded as a binary response, we use the measure of uncertainty from Eq.~\eqref{eqn:classification_uncertainty}.


\begin{figure}[t]
\centering
\includegraphics[width=\textwidth]{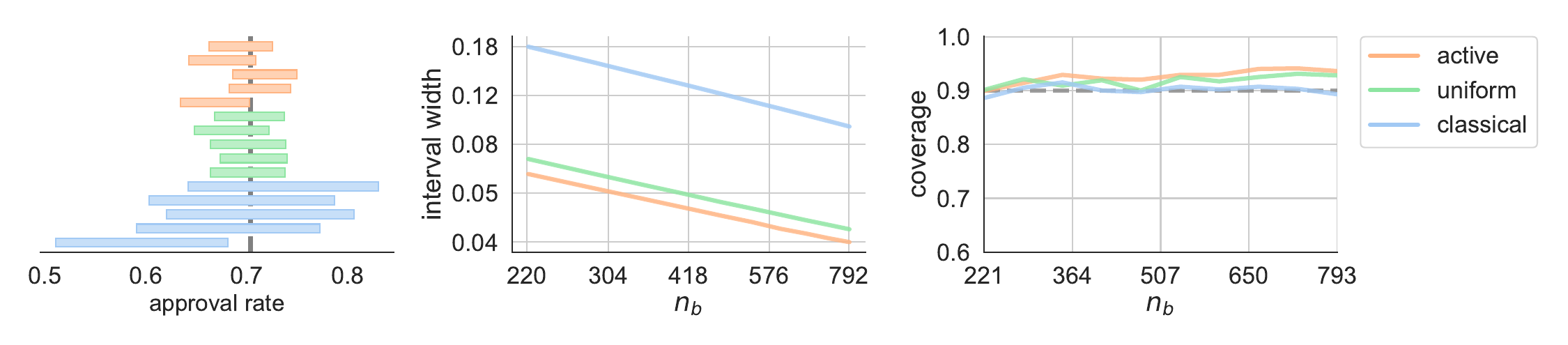}
\includegraphics[width=\textwidth]{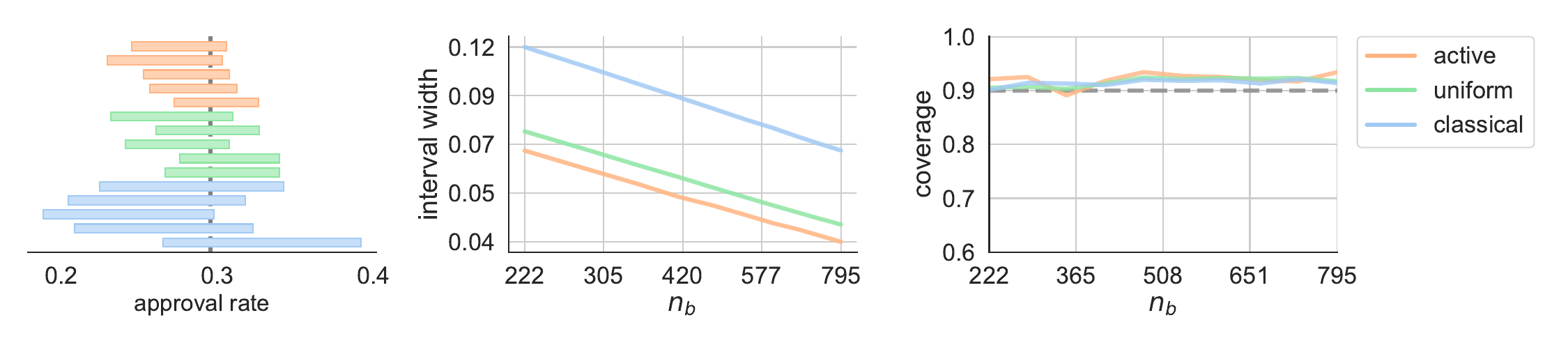}
\caption{\textbf{Post-election survey research.} Example intervals in five randomly chosen trials (left), average confidence interval width (middle), and coverage (right) for the average approval of Joe Biden's (top) and Donald Trump's (bottom) political messaging to the country following the 2020 US presidential election.}
\label{fig:pew79_batch}
\end{figure}

In Figure \ref{fig:pew79_batch} we compare active inference to the uniform (PPI) and classical baselines. All methods meet the coverage requirement.
Across different values of the budget $\nb$, active sampling reduces the confidence interval width of the uniform baseline (PPI) by a significant margin (at least $\sim 10\%$). Classical inference is highly suboptimal compared to both alternatives. In Figure \ref{fig:budgetsave} we report the percentage of budget saved due to active sampling. For estimating Biden's approval, we observe an over $85\%$ save in budget over classical inference and around $25\%$ save over the uniform baseline. For estimating Trump's approval, we observe an over $70\%$ save in budget over classical inference and around $25\%$ save over the uniform baseline.

\begin{figure}[t]
\centering
\includegraphics[width=\textwidth]{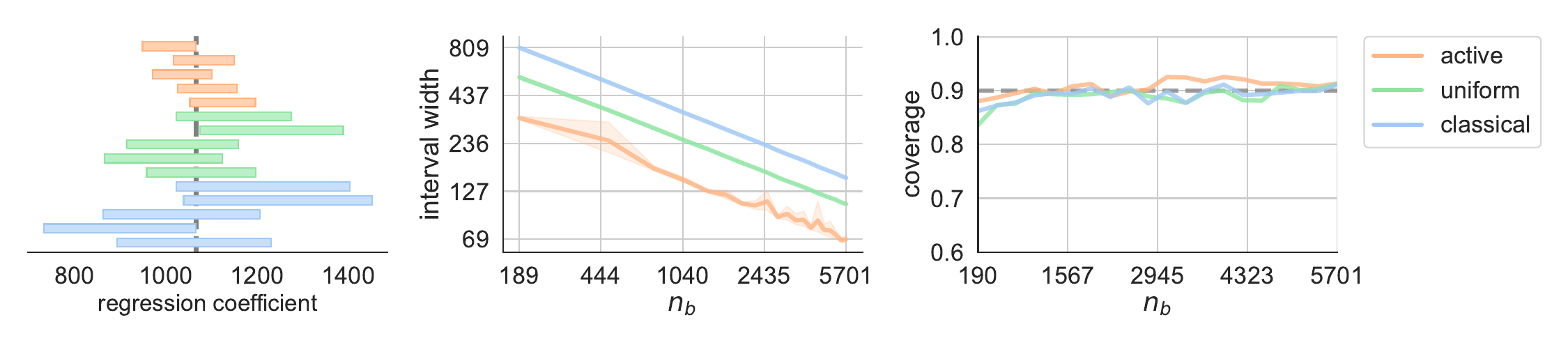}
\caption{\textbf{Census data analysis.} Example intervals in five randomly chosen trials (left), average confidence interval width (middle), and coverage (right) for the linear regression coefficient quantifying the relationship between age and income, controlling for sex, in US Census data.}
\label{fig:pums}
\end{figure}

\begin{figure}[t]
\centering
\includegraphics[width=\textwidth]{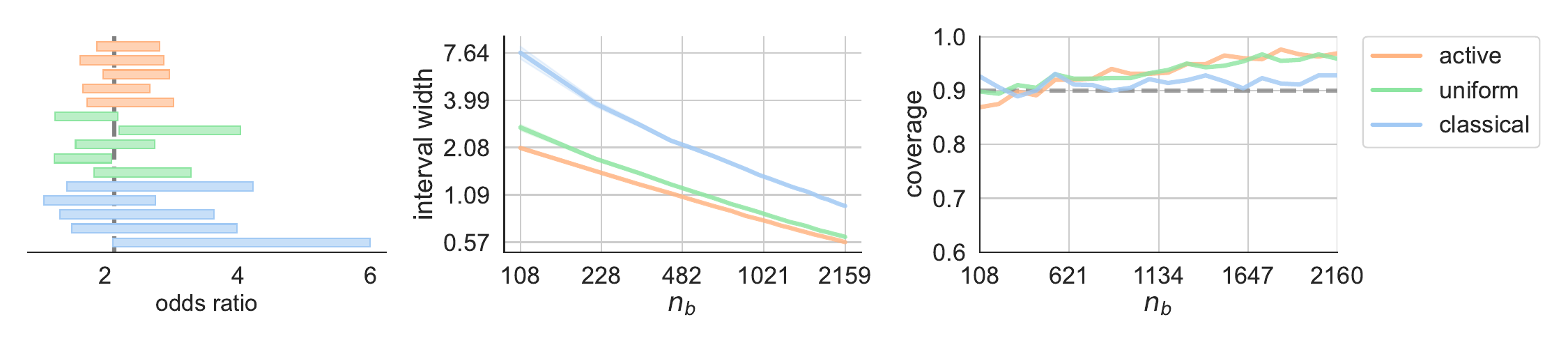}
\caption{\textbf{AlphaFold-assisted proteomics research.} Example intervals in five randomly chosen trials (left), average confidence interval width (middle), and coverage (right) for the odds ratio between phosphorylation and being part of an IDR.}
\label{fig:alphafold}
\end{figure}

\begin{figure}[t]
\centering
\includegraphics[height=150pt]{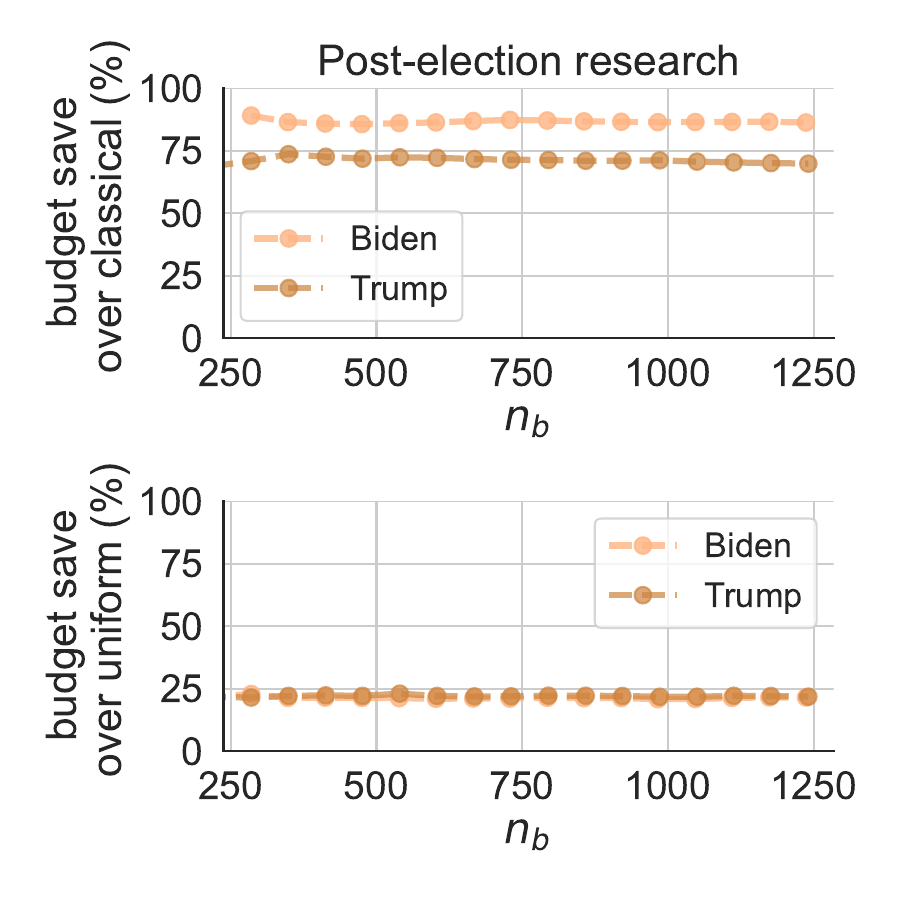}
\includegraphics[height=150pt]{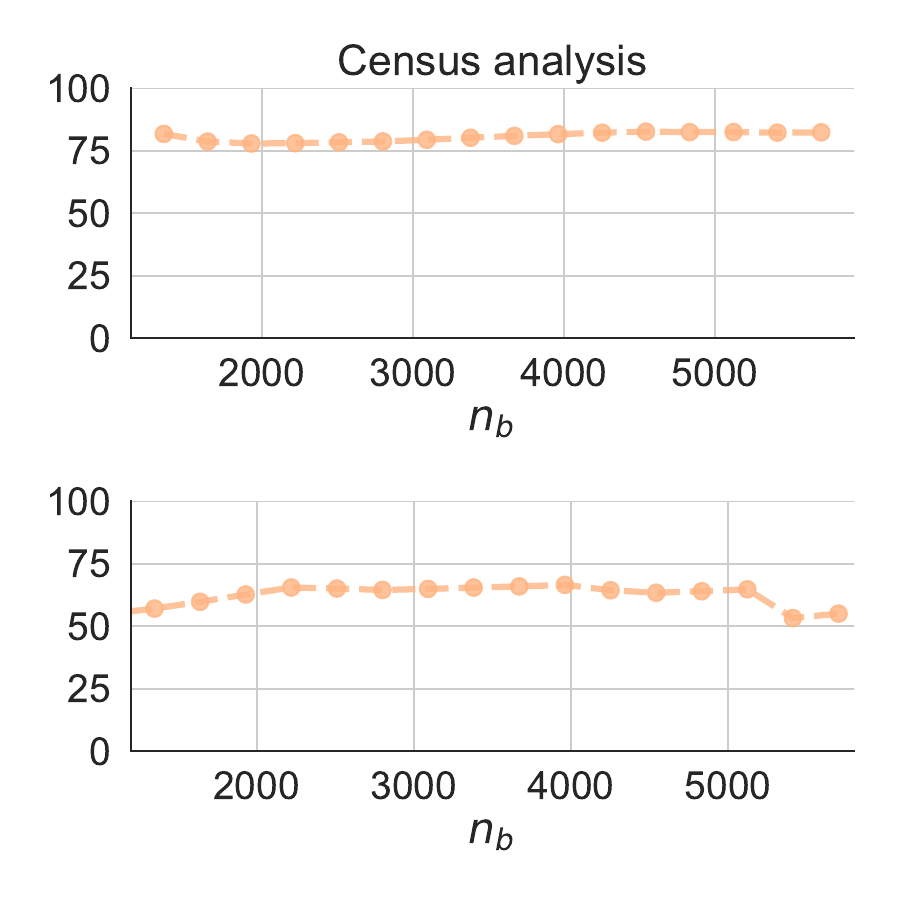}
\includegraphics[height=150pt]{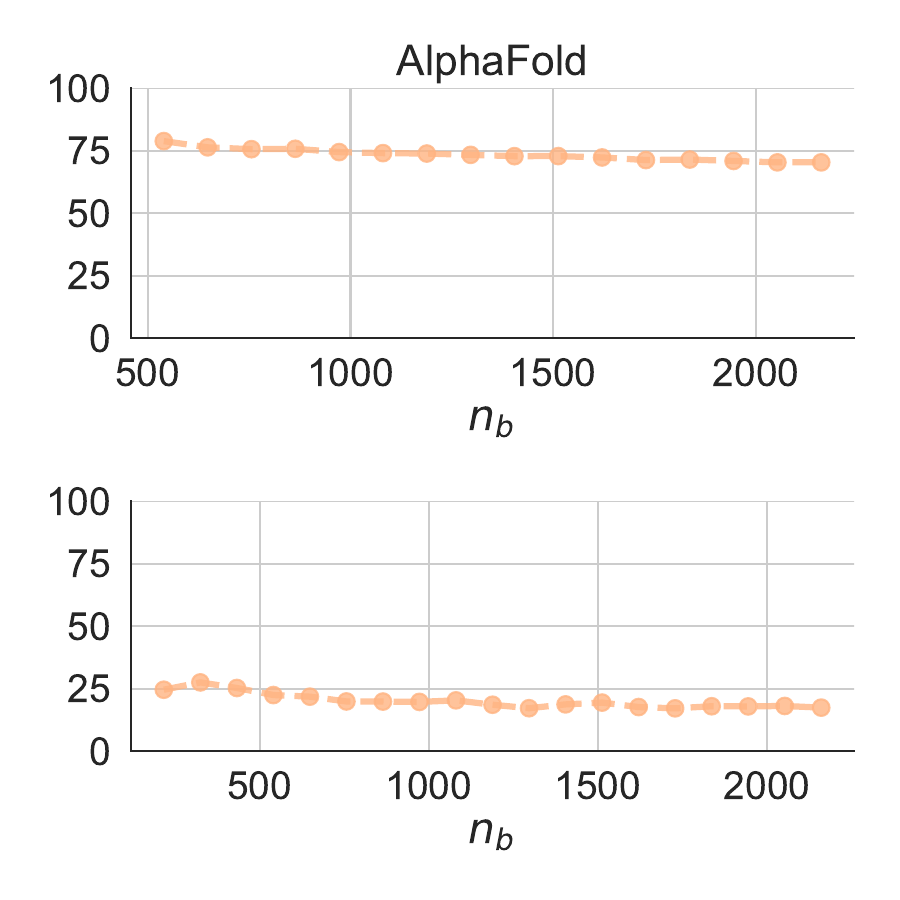}
\caption{\textbf{Save in sample budget due to active inference.} Reduction in sample size required to achieve the same confidence interval width with active inference and (top) classical inference and (bottom) uniform sampling, respectively, across the applications shown in Figures \ref{fig:pew79_batch}-\ref{fig:alphafold}.}
\label{fig:budgetsave}
\end{figure}

\subsection{Census data analysis}
\label{sec:census}

Next, we study the American Community Survey (ACS) Public Use Microdata Sample (PUMS) collected by the US Census Bureau.
ACS PUMS is an annual survey that collects information about citizenship, education, income, employment, and other factors previously contained only in the long form of the decennial census. We use the Folktables \cite{ding2021retiring} interface to download the data.
We investigate the relationship between age and income in survey data collected in California in 2019, controlling for sex.
Specifically, we target the linear regression coefficient when regressing income on age and sex (that is, its age coordinate).

Analogously to the previous application, we use half of all available data for the analysis and train an XGBoost model \cite{chen2016xgboost} of a person's income from the available demographic covariates on the other half. 
As the ground-truth value of the target $\theta^*$, we take the corresponding linear regression coefficient computed on all available data. To quantify the model's uncertainty, we use the strategy described in Section \ref{sec:basic_idea}, training a separate XGBoost model $e(\cdot)$ to predict $|f(X) - Y|$ from $X$. Then, we set the uncertainty $u(x)$ as prescribed in Eq.~\eqref{eqn:optimal_glm}, replacing $|f(X) - Y|$ by $e(X)$.

The interval widths and coverage are shown in Figure \ref{fig:pums}. As in the previous application, all methods approximately achieve the target coverage, however this time we observe more extreme gains over the uniform baseline (PPI): the interval widths almost double when going from active sampling to uniform sampling. Of course, the improvement of active inference over classical inference is even more substantial. The large gains of active sampling can also be seen in Figure \ref{fig:budgetsave}:  we save around $80\%$ of the budget over classical inference and over $60\%$ over the uniform baseline.

\subsection{AlphaFold-assisted proteomics research}

Inspired by the findings of \citet{bludau2022structural} and the subsequent analysis of \citet{angelopoulos2023prediction}, we study the odds ratio of a protein being phosphorylated, a functional property of a protein, and being part of an intrinsically disordered region (IDR), a structural property. The latter can only be obtained from knowledge about the protein structure, which can in turn be measured to a high accuracy only via expensive experimental techniques. To overcome this challenge, Bludau et al. used AlphaFold predictions~\cite{jumper2021highly} to estimate the odds ratio. AlphaFold is a machine learning model that predicts a protein's structure from its amino acid sequence.
\citet{angelopoulos2023prediction} showed that forming a classical confidence interval around the odds ratio based on AlphaFold predictions is not valid given that the predictions are imperfect. They provide a valid alternative assuming access to a small subset of proteins with true structure measurements, uniformly sampled from the larger population of proteins of interest.

We show that, by strategically choosing which protein structures to experimentally measure, active inference allows for intervals that retain validity and are tighter than intervals based on uniform sampling. Naturally, for the purpose of evaluating validity, we restrict the analysis to proteins where we have gold-standard structure measurements; we use the post-processed AlphaFold outputs made available by \citet{angelopoulos2023prediction}, which predict the IDR property based on the raw AlphaFold output.
We leverage the predictions to guide the choice of which structures to experimentally derive, subject to a budget constraint. The odds ratio we aim to estimate is defined as:
$$\theta^* = \frac{\mu_1/(1-\mu_1)}{\mu_0/(1-\mu_0)},$$
where $\mu_1 = \P(Y = 1 | X_{\mathrm{ph}} = 1)$ and $\mu_0 = \P(Y = 1 | X_{\mathrm{ph}} = 0)$; $Y$ is a binary indicator of disorder and $X_{\mathrm{ph}}$ is a binary indicator of phosphorylation.
While the odds ratio is not a solution to an M-estimation problem, it is a function of two means, $\mu_1$ and $\mu_0$ (see also \cite{angelopoulos2023prediction,angelopoulos2023ppipp}).
Confidence intervals can thus be computed by applying the delta method to the asymptotic normality result for the mean. Since $Y$ is binary, we use the measure of uncertainty from Eq.~\eqref{eqn:classification_uncertainty} to estimate $\mu_1$ and $\mu_0$. For the purpose of evaluating coverage, we take the empirical odds ratio computed on the whole dataset as the ground-truth value of $\theta^*$.

Figure \ref{fig:alphafold} shows the interval widths and coverage for the three methods, and Figure \ref{fig:budgetsave} shows the percentage of budget saved due to adaptive data collection. The gains are substantial: over $75\%$ of the budget is saved in comparison to classical inference, and around $20-25\%$ is saved in comparison to the uniform baseline (PPI). Given the cost of experimental measurement techniques in proteomics, this save in sample size would imply a massive save in cost.

\begin{figure}[t]
\centering
\includegraphics[width=1.02\textwidth]{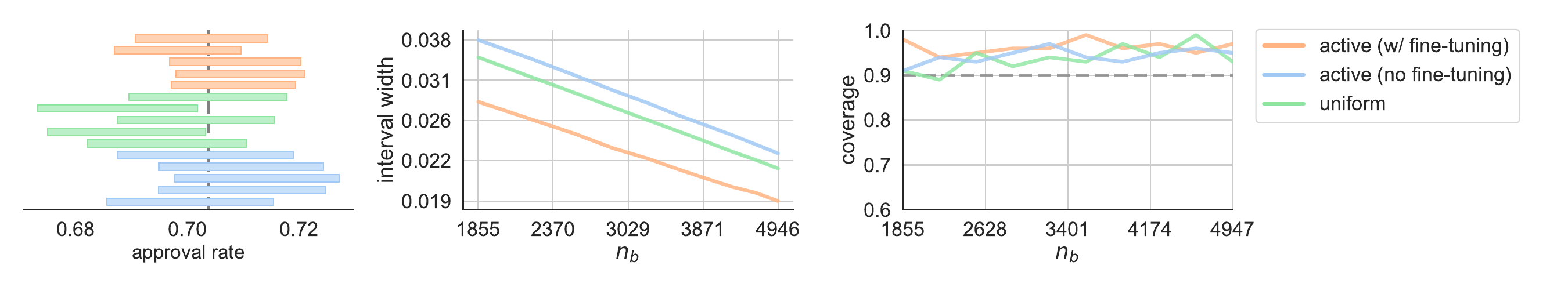}
\includegraphics[width=1.02\textwidth]{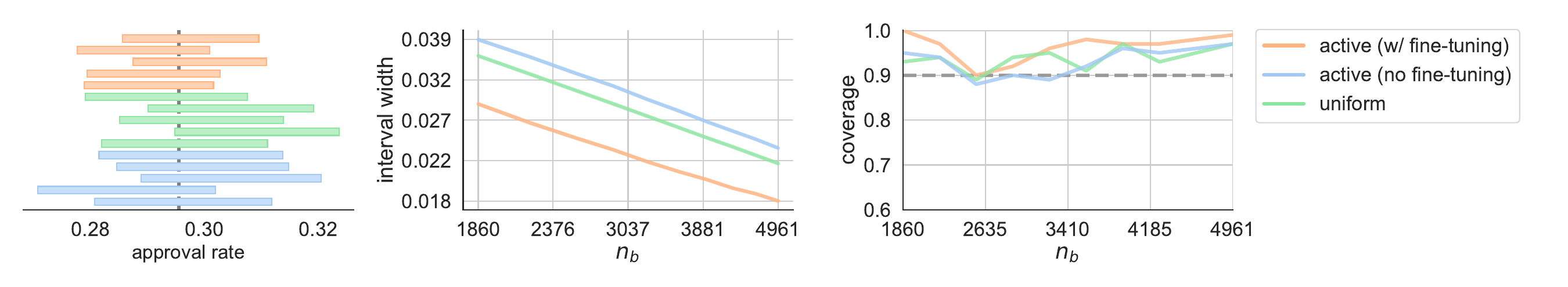}
\caption{\textbf{Post-election survey research with fine-tuning.} Example intervals in five randomly chosen trials (left), average confidence interval width (middle), and coverage (right) for the average approval of Joe Biden's (top) and Donald Trump's (bottom) political messaging to the country following the 2020 US presidential election. Active inference with no fine-tuning and inference with uniformly sampled data use the same model.}
\label{fig:pew79_finetuning}
\end{figure}

\subsection{Post-election survey research with fine-tuning}

We return to the example from Section \ref{sec:election}, this time evaluating the benefits of sequential fine-tuning. We compare active inference, with and without fine-tuning, and PPI, which relies on uniform sampling. We show that active inference with no fine-tuning can hurt compared to PPI if the former uses a poorly trained model; fine-tuning, on the other hand, remedies this issue. The predictive model may be poorly trained due to little or no historical data; sequential fine-tuning is necessary in such cases.

We train an XGBoost model on only $10$ labeled examples and use this model for active inference with no fine-tuning and PPI. The latter is similar to the former in the sense that it only replaces active with uniform sampling. Active inference with fine-tuning continues to fine-tune the model with every $B=100$ new survey responses, also updating the sampling rule via update \eqref{eqn:sequential_rule}. The uncertainty measure $u_t(x)$ is given by Eq.~\eqref{eqn:classification_uncertainty}, as before. As discussed in Section \ref{sec:sequential}, we also periodically use up the remaining budget regardless of the computed uncertainty in order to avoid underutilizing the budget (in particular, every $100n/n_b$ steps).
We fine-tune the model using the training continuation feature of XGBoost.

The interval widths and coverage are reported in Figure~\ref{fig:pew79_finetuning}. We find that fine-tuning substantially improves inferential power and retains correct coverage. In Figure \ref{fig:budgetsave_sequential} we show the save in sample size budget over active inference with no fine-tuning and inference based on uniform sampling, i.e.~PPI. For estimating Biden's approval, we observe a gain of around $40\%$ and $30\%$ relative to active inference without fine-tuning and PPI, respectively. For Trump's approval, we observe even larger gains around $45\%$ and $35\%$, respectively.

\subsection{Census data analysis with fine-tuning}

We similarly evaluate the benefits of sequential fine-tuning in the problem setting from Section \ref{sec:census}. We again compare active inference, with and without fine-tuning, and PPI, i.e., active inference with a trivial, uniform sampling rule. Recall that in Section \ref{sec:census} we trained a separate model $e$ to predict the prediction errors, which we in turn used to form the uncertainty $u(x)$ according to Eq.~\eqref{eqn:optimal_glm}. This time we fine-tune both the prediction model, $f_t$, and the error model, $e_t$.

We train initial XGBoost models $f_1$ and $e_1$ on $100$ labeled examples. We use $f_1$ for PPI and both $f_1$ and $e_1$ for active inference with no fine-tuning.  Active inference with fine-tuning continues to fine-tune the two models with every $B=1000$ new survey responses, also updating the model uncertainty via update \eqref{eqn:sequential_rule}. We fine-tune the models using the training continuation feature of XGBoost. We compute $u_t$ from $e_t$ based on Eq.~\eqref{eqn:optimal_glm}. As discussed earlier, we also periodically use up the remaining budget regardless of the computed uncertainty in order to avoid underutilizing the budget (in particular, every $500n/n_b$ steps).

\begin{figure}[t]
\centering
\includegraphics[width=1.02\textwidth]{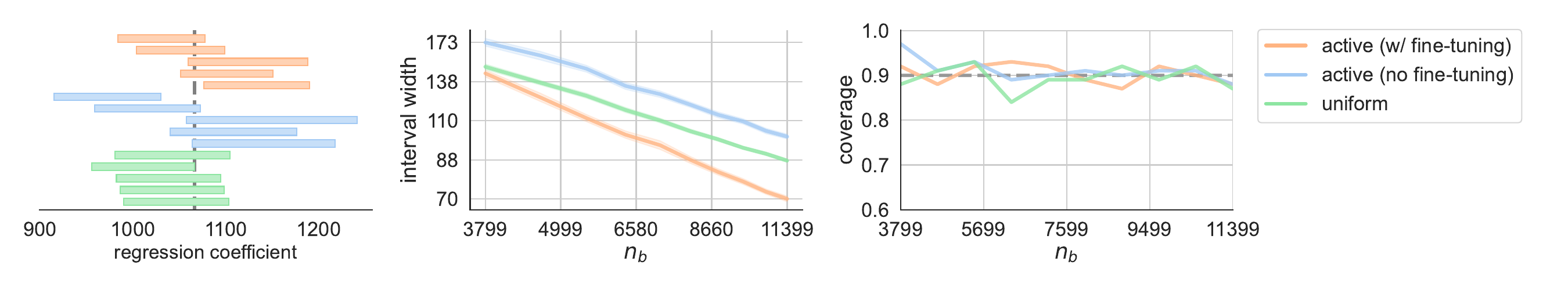}
\caption{\textbf{Census data analysis with fine-tuning.} Example intervals in five randomly chosen trials (left), average confidence interval width (middle), and coverage (right) for the linear regression coefficient quantifying the relationship between age and income, controlling for sex, in US Census data. Active inference with no fine-tuning and inference with uniformly sampled data use the same model.}
\label{fig:census_finetuning}
\end{figure}

\begin{figure}[t]
\centering
\includegraphics[height=150pt]{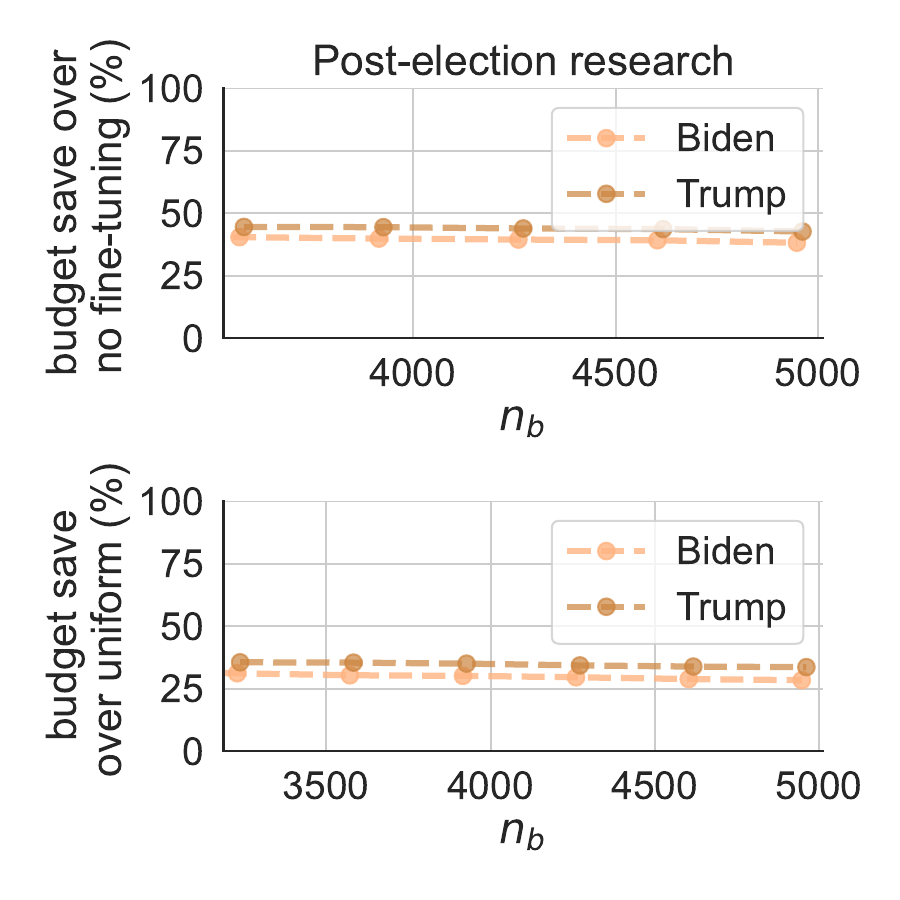}
\includegraphics[height=150pt]{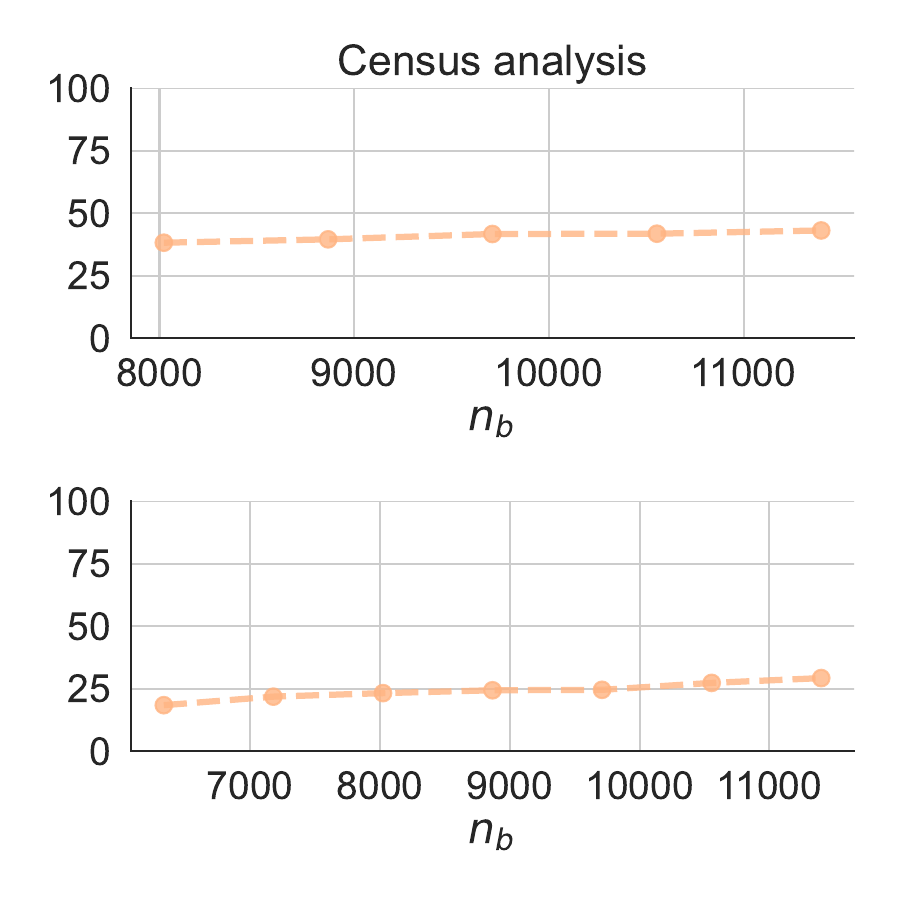}
\caption{\textbf{Save in sample size budget due to fine-tuning.} Reduction in sample size required to achieve the same confidence interval width with active inference with fine-tuning and (top) active inference with no fine-tuning and (bottom) the uniform baseline (PPI), respectively, in the applications shown in Figure \ref{fig:pew79_finetuning} and Figure~\ref{fig:census_finetuning}.}
\label{fig:budgetsave_sequential}
\end{figure}

We show the interval widths and coverage in Figure~\ref{fig:census_finetuning}. We see that the gains of fine-tuning are significant and increase as $n_b$ increases. In Figure \ref{fig:budgetsave_sequential} we show the save in sample size budget. Fine-tuning saves around $32-40\%$ over the baseline with no fine-tuning and around $20-30\%$ over the uniform baseline. Moreover, the save increases as the sample budget grows because the prediction problem is difficult and the model's performance keeps improving even after $10000$ training examples.

\section*{Acknowledgements}

We thank Lihua Lei, Jann Spiess, and Stefan Wager for many insightful comments and pointers to relevant work. T.Z.~was supported by Stanford Data Science through the Fellowship program. E.J.C.~was supported by the Office of Naval Research grant N00014-24-1-2305, the National Science Foundation grant DMS-2032014, the Simons Foundation under award 814641, and the ARO grant 2003514594.

\bibliographystyle{plainnat}
\bibliography{refs}

\begin{thebibliography}{61}
\providecommand{\natexlab}[1]{#1}
\providecommand{\url}[1]{\texttt{#1}}
\expandafter\ifx\csname urlstyle\endcsname\relax
  \providecommand{\doi}[1]{doi: #1}\else
  \providecommand{\doi}{doi: \begingroup \urlstyle{rm}\Url}\fi

\bibitem[Angelopoulos et~al.(2023{\natexlab{a}})Angelopoulos, Bates, Fannjiang, Jordan, and Zrnic]{angelopoulos2023prediction}
Anastasios~N Angelopoulos, Stephen Bates, Clara Fannjiang, Michael~I Jordan, and Tijana Zrnic.
\newblock Prediction-powered inference.
\newblock \emph{Science}, 382\penalty0 (6671):\penalty0 669--674, 2023{\natexlab{a}}.

\bibitem[Angelopoulos et~al.(2023{\natexlab{b}})Angelopoulos, Bates, Fannjiang, Jordan, and Zrnic]{ppidata}
Anastasios~N Angelopoulos, Stephen Bates, Clara Fannjiang, Michael~I Jordan, and Tijana Zrnic.
\newblock Prediction-powered inference: Data sets, 2023{\natexlab{b}}.
\newblock URL \url{https://doi.org/10.5281/zenodo.8397451}.

\bibitem[Angelopoulos et~al.(2023{\natexlab{c}})Angelopoulos, Duchi, and Zrnic]{angelopoulos2023ppipp}
Anastasios~N Angelopoulos, John~C Duchi, and Tijana Zrnic.
\newblock {PPI}++: Efficient prediction-powered inference.
\newblock \emph{arXiv preprint arXiv:2311.01453}, 2023{\natexlab{c}}.

\bibitem[Ash et~al.(2019)Ash, Zhang, Krishnamurthy, Langford, and Agarwal]{ash2019deep}
Jordan~T Ash, Chicheng Zhang, Akshay Krishnamurthy, John Langford, and Alekh Agarwal.
\newblock Deep batch active learning by diverse, uncertain gradient lower bounds.
\newblock \emph{arXiv preprint arXiv:1906.03671}, 2019.

\bibitem[Azriel et~al.(2022)Azriel, Brown, Sklar, Berk, Buja, and Zhao]{azriel2022semi}
David Azriel, Lawrence~D Brown, Michael Sklar, Richard Berk, Andreas Buja, and Linda Zhao.
\newblock Semi-supervised linear regression.
\newblock \emph{Journal of the American Statistical Association}, 117\penalty0 (540):\penalty0 2238--2251, 2022.

\bibitem[Balcan et~al.(2006)Balcan, Beygelzimer, and Langford]{balcan2006agnostic}
Maria-Florina Balcan, Alina Beygelzimer, and John Langford.
\newblock Agnostic active learning.
\newblock In \emph{Proceedings of the 23rd international conference on Machine learning}, pages 65--72, 2006.

\bibitem[Balcan et~al.(2014)Balcan, Daniely, Mehta, Urner, and Vazirani]{balcan2014learning}
Maria-Florina Balcan, Amit Daniely, Ruta Mehta, Ruth Urner, and Vijay~V Vazirani.
\newblock Learning economic parameters from revealed preferences.
\newblock In \emph{Web and Internet Economics: 10th International Conference, WINE 2014, Beijing, China, December 14-17, 2014. Proceedings 10}, pages 338--353. Springer, 2014.

\bibitem[Bhattacharya and Dupas(2012)]{bhattacharya2012inferring}
Debopam Bhattacharya and Pascaline Dupas.
\newblock Inferring welfare maximizing treatment assignment under budget constraints.
\newblock \emph{Journal of Econometrics}, 167\penalty0 (1):\penalty0 168--196, 2012.

\bibitem[Bludau et~al.(2022)Bludau, Willems, Zeng, Strauss, Hansen, Tanzer, Karayel, Schulman, and Mann]{bludau2022structural}
Isabell Bludau, Sander Willems, Wen-Feng Zeng, Maximilian~T Strauss, Fynn~M Hansen, Maria~C Tanzer, Ozge Karayel, Brenda~A Schulman, and Matthias Mann.
\newblock The structural context of posttranslational modifications at a proteome-wide scale.
\newblock \emph{PLoS biology}, 20\penalty0 (5):\penalty0 e3001636, 2022.

\bibitem[Chandak et~al.(2023)Chandak, Shankar, Syrgkanis, and Brunskill]{chandak2023adaptive}
Yash Chandak, Shiv Shankar, Vasilis Syrgkanis, and Emma Brunskill.
\newblock Adaptive instrument design for indirect experiments.
\newblock \emph{arXiv preprint arXiv:2312.02438}, 2023.

\bibitem[Chen and Guestrin(2016)]{chen2016xgboost}
Tianqi Chen and Carlos Guestrin.
\newblock Xgboost: A scalable tree boosting system.
\newblock In \emph{Proceedings of the 22nd acm sigkdd international conference on knowledge discovery and data mining}, pages 785--794, 2016.

\bibitem[Cheng et~al.(2022)Cheng, Asi, and Duchi]{cheng2022many}
Chen Cheng, Hilal Asi, and John Duchi.
\newblock How many labelers do you have? a closer look at gold-standard labels.
\newblock \emph{arXiv preprint arXiv:2206.12041}, 2022.

\bibitem[Chernozhukov et~al.(2018)Chernozhukov, Chetverikov, Demirer, Duflo, Hansen, Newey, and Robins]{chernozhukov2018double}
Victor Chernozhukov, Denis Chetverikov, Mert Demirer, Esther Duflo, Christian Hansen, Whitney Newey, and James Robins.
\newblock Double/debiased machine learning for treatment and structural parameters, 2018.

\bibitem[Cook et~al.(2023)Cook, Mishler, and Ramdas]{cook2023semiparametric}
Thomas Cook, Alan Mishler, and Aaditya Ramdas.
\newblock Semiparametric efficient inference in adaptive experiments.
\newblock \emph{arXiv preprint arXiv:2311.18274}, 2023.

\bibitem[Ding et~al.(2021)Ding, Hardt, Miller, and Schmidt]{ding2021retiring}
Frances Ding, Moritz Hardt, John Miller, and Ludwig Schmidt.
\newblock Retiring adult: New datasets for fair machine learning.
\newblock \emph{Advances in neural information processing systems}, 34:\penalty0 6478--6490, 2021.

\bibitem[Durrett(2019)]{durrett2019probability}
Rick Durrett.
\newblock \emph{Probability: theory and examples}, volume~49.
\newblock Cambridge university press, 2019.

\bibitem[Dvoretzky(1972)]{dvoretzky1972asymptotic}
Aryeh Dvoretzky.
\newblock Asymptotic normality for sums of dependent random variables.
\newblock In \emph{Proceedings of the Sixth Berkeley Symposium on Mathematical Statistics and Probability, Volume 2: Probability Theory}, volume~6, pages 513--536. University of California Press, 1972.

\bibitem[Gal et~al.(2017)Gal, Islam, and Ghahramani]{gal2017deep}
Yarin Gal, Riashat Islam, and Zoubin Ghahramani.
\newblock Deep {B}ayesian active learning with image data.
\newblock In \emph{International conference on machine learning}, pages 1183--1192. PMLR, 2017.

\bibitem[Gan and Liang(2023)]{gan2023prediction}
Feng Gan and Wanfeng Liang.
\newblock Prediction de-correlated inference.
\newblock \emph{arXiv preprint arXiv:2312.06478}, 2023.

\bibitem[Hadad et~al.(2021)Hadad, Hirshberg, Zhan, Wager, and Athey]{hadad2021confidence}
Vitor Hadad, David~A Hirshberg, Ruohan Zhan, Stefan Wager, and Susan Athey.
\newblock Confidence intervals for policy evaluation in adaptive experiments.
\newblock \emph{Proceedings of the national academy of sciences}, 118\penalty0 (15):\penalty0 e2014602118, 2021.

\bibitem[Hahn et~al.(2011)Hahn, Hirano, and Karlan]{hahn2011adaptive}
Jinyong Hahn, Keisuke Hirano, and Dean Karlan.
\newblock Adaptive experimental design using the propensity score.
\newblock \emph{Journal of Business \& Economic Statistics}, 29\penalty0 (1):\penalty0 96--108, 2011.

\bibitem[Hanneke et~al.(2014)]{hanneke2014theory}
Steve Hanneke et~al.
\newblock Theory of disagreement-based active learning.
\newblock \emph{Foundations and Trends{\textregistered} in Machine Learning}, 7\penalty0 (2-3):\penalty0 131--309, 2014.

\bibitem[Hu and Rosenberger(2006)]{hu2006theory}
Feifang Hu and William~F Rosenberger.
\newblock \emph{The theory of response-adaptive randomization in clinical trials}.
\newblock John Wiley \& Sons, 2006.

\bibitem[Jean et~al.(2016)Jean, Burke, Xie, Davis, Lobell, and Ermon]{jean2016combining}
Neal Jean, Marshall Burke, Michael Xie, W~Matthew Davis, David~B Lobell, and Stefano Ermon.
\newblock Combining satellite imagery and machine learning to predict poverty.
\newblock \emph{Science}, 353\penalty0 (6301):\penalty0 790--794, 2016.

\bibitem[Joshi et~al.(2009)Joshi, Porikli, and Papanikolopoulos]{joshi2009multi}
Ajay~J Joshi, Fatih Porikli, and Nikolaos Papanikolopoulos.
\newblock Multi-class active learning for image classification.
\newblock In \emph{2009 ieee conference on computer vision and pattern recognition}, pages 2372--2379. IEEE, 2009.

\bibitem[Jumper et~al.(2021)Jumper, Evans, Pritzel, Green, Figurnov, Ronneberger, Tunyasuvunakool, Bates, {\v{Z}}{\'\i}dek, Potapenko, et~al.]{jumper2021highly}
John Jumper, Richard Evans, Alexander Pritzel, Tim Green, Michael Figurnov, Olaf Ronneberger, Kathryn Tunyasuvunakool, Russ Bates, Augustin {\v{Z}}{\'\i}dek, Anna Potapenko, et~al.
\newblock Highly accurate protein structure prediction with alphafold.
\newblock \emph{Nature}, 596\penalty0 (7873):\penalty0 583--589, 2021.

\bibitem[Kalton(2020)]{kalton2020introduction}
Graham Kalton.
\newblock \emph{Introduction to survey sampling}.
\newblock Number~35. Sage Publications, 2020.

\bibitem[Kasy and Sautmann(2021)]{kasy2021adaptive}
Maximilian Kasy and Anja Sautmann.
\newblock Adaptive treatment assignment in experiments for policy choice.
\newblock \emph{Econometrica}, 89\penalty0 (1):\penalty0 113--132, 2021.

\bibitem[Kato et~al.(2020)Kato, Ishihara, Honda, and Narita]{kato2020efficient}
Masahiro Kato, Takuya Ishihara, Junya Honda, and Yusuke Narita.
\newblock Efficient adaptive experimental design for average treatment effect estimation.
\newblock \emph{arXiv preprint arXiv:2002.05308}, 2020.

\bibitem[Khan et~al.(2015)Khan, Reddy, and Rao]{khan2015designing}
Mohammad~GM Khan, Karuna~G Reddy, and Dinesh~K Rao.
\newblock Designing stratified sampling in economic and business surveys.
\newblock \emph{Journal of applied statistics}, 42\penalty0 (10):\penalty0 2080--2099, 2015.

\bibitem[Lai and Robbins(1985)]{lai1985asymptotically}
Tze~Leung Lai and Herbert Robbins.
\newblock Asymptotically efficient adaptive allocation rules.
\newblock \emph{Advances in applied mathematics}, 6\penalty0 (1):\penalty0 4--22, 1985.

\bibitem[List et~al.(2011)List, Sadoff, and Wagner]{list2011so}
John~A List, Sally Sadoff, and Mathis Wagner.
\newblock So you want to run an experiment, now what? some simple rules of thumb for optimal experimental design.
\newblock \emph{Experimental Economics}, 14:\penalty0 439--457, 2011.

\bibitem[Miao et~al.(2023)Miao, Miao, Wu, Zhao, and Lu]{miao2023assumption}
Jiacheng Miao, Xinran Miao, Yixuan Wu, Jiwei Zhao, and Qiongshi Lu.
\newblock Assumption-lean and data-adaptive post-prediction inference.
\newblock \emph{arXiv preprint arXiv:2311.14220}, 2023.

\bibitem[Motwani and Witten(2023)]{motwani2023valid}
Keshav Motwani and Daniela Witten.
\newblock Valid inference after prediction.
\newblock \emph{arXiv preprint arXiv:2306.13746}, 2023.

\bibitem[Nassiuma(2001)]{nassiuma2001survey}
Dankit~K Nassiuma.
\newblock Survey sampling: Theory and methods, 2001.

\bibitem[Orabona and Jun(2023)]{orabona2023tight}
Francesco Orabona and Kwang-Sung Jun.
\newblock Tight concentrations and confidence sequences from the regret of universal portfolio.
\newblock \emph{IEEE Transactions on Information Theory}, 2023.

\bibitem[Owen(2013)]{owen_mcbook}
Art~B. Owen.
\newblock \emph{Monte Carlo theory, methods and examples}.
\newblock \url{https://artowen.su.domains/mc/}, 2013.

\bibitem[Pew(2020)]{atp79}
Pew.
\newblock American trends panel ({ATP}) wave 79, 2020.
\newblock URL \url{https://www.pewresearch.org/science/dataset/american-trends-panel-wave-79/}.

\bibitem[Ren et~al.(2021)Ren, Xiao, Chang, Huang, Li, Gupta, Chen, and Wang]{ren2021survey}
Pengzhen Ren, Yun Xiao, Xiaojun Chang, Po-Yao Huang, Zhihui Li, Brij~B Gupta, Xiaojiang Chen, and Xin Wang.
\newblock A survey of deep active learning.
\newblock \emph{ACM computing surveys (CSUR)}, 54\penalty0 (9):\penalty0 1--40, 2021.

\bibitem[Robbins(1952)]{robbins1952some}
Herbert Robbins.
\newblock Some aspects of the sequential design of experiments.
\newblock 1952.

\bibitem[Robins and Rotnitzky(1995)]{robins1995semiparametric}
James~M Robins and Andrea Rotnitzky.
\newblock Semiparametric efficiency in multivariate regression models with missing data.
\newblock \emph{Journal of the American Statistical Association}, 90\penalty0 (429):\penalty0 122--129, 1995.

\bibitem[Robins et~al.(1994)Robins, Rotnitzky, and Zhao]{robins1994estimation}
James~M Robins, Andrea Rotnitzky, and Lue~Ping Zhao.
\newblock Estimation of regression coefficients when some regressors are not always observed.
\newblock \emph{Journal of the American statistical Association}, 89\penalty0 (427):\penalty0 846--866, 1994.

\bibitem[Rolf et~al.(2021)Rolf, Proctor, Carleton, Bolliger, Shankar, Ishihara, Recht, and Hsiang]{rolf2021generalizable}
Esther Rolf, Jonathan Proctor, Tamma Carleton, Ian Bolliger, Vaishaal Shankar, Miyabi Ishihara, Benjamin Recht, and Solomon Hsiang.
\newblock A generalizable and accessible approach to machine learning with global satellite imagery.
\newblock \emph{Nature communications}, 12\penalty0 (1):\penalty0 4392, 2021.

\bibitem[Rubin(1987)]{rubin1987multiple}
D~Rubin.
\newblock Multiple imputation for nonresponse in surveys.
\newblock \emph{Wiley Series in Probability and Statistics}, page~1, 1987.

\bibitem[Rubin(1976)]{rubin1976inference}
Donald~B Rubin.
\newblock Inference and missing data.
\newblock \emph{Biometrika}, 63\penalty0 (3):\penalty0 581--592, 1976.

\bibitem[Rubin(1996)]{rubin1996multiple}
Donald~B Rubin.
\newblock Multiple imputation after 18+ years.
\newblock \emph{Journal of the American statistical Association}, 91\penalty0 (434):\penalty0 473--489, 1996.

\bibitem[S{\"a}rndal(1980)]{sarndal1980pi}
Carl~Erik S{\"a}rndal.
\newblock On $\pi$-inverse weighting versus best linear unbiased weighting in probability sampling.
\newblock \emph{Biometrika}, 67\penalty0 (3):\penalty0 639--650, 1980.

\bibitem[S{\"a}rndal et~al.(2003)S{\"a}rndal, Swensson, and Wretman]{sarndal2003model}
Carl-Erik S{\"a}rndal, Bengt Swensson, and Jan Wretman.
\newblock \emph{Model assisted survey sampling}.
\newblock Springer Science \& Business Media, 2003.

\bibitem[Schohn and Cohn(2000)]{schohn2000less}
Greg Schohn and David Cohn.
\newblock Less is more: Active learning with support vector machines.
\newblock In \emph{ICML}, volume~2, page~6, 2000.

\bibitem[Settles(2009)]{settles2009active}
Burr Settles.
\newblock Active learning literature survey.
\newblock \emph{Department of Computer Sciences, University of Wisconsin-Madison}, 2009.

\bibitem[Tong and Koller(2001)]{tong2001support}
Simon Tong and Daphne Koller.
\newblock Support vector machine active learning with applications to text classification.
\newblock \emph{Journal of machine learning research}, 2\penalty0 (Nov):\penalty0 45--66, 2001.

\bibitem[Van~der Vaart(2000)]{van2000asymptotic}
Aad~W Van~der Vaart.
\newblock \emph{Asymptotic statistics}, volume~3.
\newblock Cambridge university press, 2000.

\bibitem[Vishwakarma et~al.(2023)Vishwakarma, Lin, Sala, and Korlakai~Vinayak]{vishwakarma2022good}
Harit Vishwakarma, Heguang Lin, Frederic Sala, and Ramya Korlakai~Vinayak.
\newblock Promises and pitfalls of threshold-based auto-labeling.
\newblock \emph{Advances in Neural Information Processing Systems}, 36, 2023.

\bibitem[Waudby-Smith and Ramdas(2024)]{waudby2020estimating}
Ian Waudby-Smith and Aaditya Ramdas.
\newblock Estimating means of bounded random variables by betting.
\newblock \emph{Journal of the Royal Statistical Society Series B: Statistical Methodology}, 86\penalty0 (1):\penalty0 1--27, 2024.

\bibitem[Xie et~al.(2016)Xie, Jean, Burke, Lobell, and Ermon]{xie2016transfer}
Michael Xie, Neal Jean, Marshall Burke, David Lobell, and Stefano Ermon.
\newblock Transfer learning from deep features for remote sensing and poverty mapping.
\newblock In \emph{Proceedings of the AAAI conference on artificial intelligence}, volume~30, 2016.

\bibitem[Zdun(2022)]{uselections}
Matt Zdun.
\newblock Machine politics: How {A}merica casts and counts its votes.
\newblock \emph{Reuters}, 2022.

\bibitem[Zhang et~al.(2019)Zhang, Brown, and Cai]{zhang2019semi}
Anru Zhang, Lawrence~D Brown, and T~Tony Cai.
\newblock Semi-supervised inference: General theory and estimation of means.
\newblock \emph{Annals of Statistics}, 47\penalty0 (5):\penalty0 2538--2566, 2019.

\bibitem[Zhang et~al.(2023)Zhang, Cammarata, Squires, Sapsis, and Uhler]{zhang2023active}
Jiaqi Zhang, Louis Cammarata, Chandler Squires, Themistoklis~P Sapsis, and Caroline Uhler.
\newblock Active learning for optimal intervention design in causal models.
\newblock \emph{Nature Machine Intelligence}, pages 1--10, 2023.

\bibitem[Zhang et~al.(2021)Zhang, Janson, and Murphy]{zhang2021statistical}
Kelly Zhang, Lucas Janson, and Susan Murphy.
\newblock Statistical inference with {M}-estimators on adaptively collected data.
\newblock \emph{Advances in neural information processing systems}, 34:\penalty0 7460--7471, 2021.

\bibitem[Zhang and Bradic(2022)]{zhang2022high}
Yuqian Zhang and Jelena Bradic.
\newblock High-dimensional semi-supervised learning: in search of optimal inference of the mean.
\newblock \emph{Biometrika}, 109\penalty0 (2):\penalty0 387--403, 2022.

\bibitem[Zrnic and Cand{\`e}s(2024)]{zrnic2023cross}
Tijana Zrnic and Emmanuel~J Cand{\`e}s.
\newblock Cross-prediction-powered inference.
\newblock \emph{Proceedings of the National Academy of Sciences}, 121\penalty0 (15):\penalty0 e2322083121, 2024.

\end{thebibliography}

\newpage
\appendix 

\section{Proofs}

\subsection{Proof of Proposition \ref{prop:mean_batch_inference}}

Recall that $\xi_i \sim\mathrm{Bern}(\pi_{\hat\eta}(X_i))$. For any $\eta\in \cH$, we define
\begin{equation}
\label{eq:xi_eta}
\xi_i^{\eta} = \one\{ \pi_{\eta}(X_i) \leq \pi_{\hat\eta}(X_i)  \}\xi_i (1-\xi_i^{\leq}) + \one\{ \pi_{\eta}(X_i) > \pi_{\hat\eta}(X_i) \}(\xi_i + (1-\xi_i)\xi_i^{>}),
\end{equation}
where $\xi_i^{\leq} \sim \mathrm{Bern}( \frac{\pi_{\hat\eta}(X_i) - \pi_{\eta}(X_i)}{\pi_{\hat\eta}(X_i)})$ and $\xi_i^{>} \sim \mathrm{Bern}(\frac{\pi_{\eta}(X_i) - \pi_{\hat\eta}(X_i)}{1-\pi_{\hat\eta}(X_i)})$ are drawn independently of $\xi_i$. This definition couples $\xi_i^{\eta^*}$ with $\xi_i$, while ensuring that $\xi_i^{\eta^*}\sim \mathrm{Bern}(\pi_{\eta^*}(X_i))$.
Let
$$\hat\theta^{\eta^*} = \frac 1 n \sum_{i=1}^n \left(f(X_i) + (Y_i - f(X_i))\frac{\xi^{\eta^*}_i}{\pi_{\eta^*}(X_i)} \right).$$
By the central limit theorem, we know that
\begin{equation}
\label{eqn:clt_etastar}
\sqrt{n}(\hat\theta^{\eta^*} - \theta^*)\cd \cN(0,\sigma_*^2),
\end{equation}
where $\sigma_*^2 = \Var\left(f(X)  + (Y - f(X)) \frac{\xi^{\eta^*}}{\pi_{\eta^*}(X)} \right)$.
On the other hand, we have
\begin{align*}
\sqrt{n}(\hat\theta^{\hat \eta} - \theta^*) &= \sqrt{n} (\hat\theta^{\eta^*} - \theta^* ) + \sqrt{n} ( \hat\theta^{\hat \eta} - \hat\theta^{\eta^*}).
\end{align*}
For any $\epsilon>0$, we have $\P(|\sqrt{n} ( \hat\theta^{\hat \eta} - \hat\theta^{\eta^*})|\geq \epsilon) \leq \P(\hat\eta\neq \eta^*)\to 0$; therefore, $\sqrt{n} ( \hat\theta^{\hat \eta} - \hat\theta^{\eta^*}) \stackrel{p}{\to} 0$. Putting this fact together with Eq.~\eqref{eqn:clt_etastar}, we conclude that
$\sqrt{n}(\hat\theta^{\hat \eta} - \theta^*) \cd \cN(0,\sigma_*^2)$ by Slutsky's theorem.

\subsection{Proof of Theorem \ref{thm:m_est_batch}}

The proof follows a similar argument as the classical proof of asymptotic normality for M-estimation; see \citep[Thm.~5.23]{van2000asymptotic}. A similar proof is also given for the prediction-powered estimator \cite{angelopoulos2023ppipp}, which is closely related to our active inference estimator. The main difference between our proof and the classical proof is that $\hat\eta$ is tuned in a data-adaptive fashion, so the increments in the empirical loss $L^{\pi_{\hat\eta}}(\theta)$ are not independent. We begin by formally stating the required smoothness assumption.

\begin{assumption}[Smoothness]
\label{ass:smooth_loss}
The loss $\ell$ is smooth if:
\begin{itemize}
\item $\ell_\theta(x,y)$ is differentiable at  $\theta^*$ for all $(x,y)$;
\item $\ell_\theta$ is locally Lipschitz around $\theta^*$: there is a neighborhood of $\theta^*$ such that $\ell_\theta(x,y)$ is $C(x,y)$-Lipschitz and $\ell_\theta(x,f(x))$ is $C(x)$-Lipschitz in $\theta$, where $\E[C(X,Y)^2] < \infty, \E[C(X)^2] < \infty$;
\item $L(\theta) = \E[\ell_\theta(X,Y)]$ and $L^f(\theta) = \E[\ell_\theta(X,f(X))]$ have Hessians, and $H_{\theta^*} = \nabla^2 L(\theta^*)\succ 0$.
\end{itemize}
\end{assumption}

Using the same definition of $\xi^{\eta}_i$ as in Eq. \eqref{eq:xi_eta}, let $L_{\theta,i}^\eta = \ell_\theta(X_i,f(X_i)) + \left(\ell_\theta(X_i,Y_i) - \ell_\theta(X_i, f(X_i)) \right) \frac{\xi^\eta_i}{\pi_{\eta}(X_i)}$. We define $\nabla L_{\theta,i}^\eta$ analogously, replacing the losses with their gradients. Given a function $g$, let
\begin{align*}
\GG_n[g(L_{\theta}^\eta)] &:= \frac{1}{\sqrt{n}} \sum_{i=1}^n \left(g(L_{\theta,i}^\eta) - \E[g(L_{\theta,i}^\eta)]\right); \quad 
\E_n[g(L_{\theta}^\eta)] := \frac{1}{n} \sum_{i=1}^n g(L_{\theta,i}^\eta).
\end{align*}
We similarly use $\GG_n[g(\nabla L_{\theta}^\eta)]$, $\E_n[g(\nabla L_{\theta}^\eta)]$, etc.
Notice that $\E_n[L_\theta^{\hat \eta}] = L^{\pi_{\hat\eta}}(\theta)$.

By the differentiability and local Lipschitzness of the loss, for any $h_n = O_P(1)$ we have
$$\GG_n[\sqrt{n}(L_{\theta^* + h_n/\sqrt{n}}^{\eta^*} - L_{\theta^*}^{\eta^*}) - h_n^\top \nabla L_{\theta^*}^{\eta^*}] \stackrel{p}{\to} 0.$$
By definition, this is equivalent to
\begin{align*}
n\E_n[L_{\theta^* + h_n/\sqrt{n}}^{\eta^*} - L_{\theta^*}^{\eta^*}] &= n(L(\theta^* + h_n/\sqrt{n}) - L(\theta^*)) + h_n^\top \GG_n [\nabla L_{\theta^*}^{\eta^*}] + o_P(1),
\end{align*}
where $L(\theta) = \E[\ell_\theta(X,Y)]$ is the population loss.
A second-order Taylor expansion now implies
$$n\E_n[L_{\theta^* + h_n/\sqrt{n}}^{\eta^*} - L_{\theta^*}^{\eta^*}]  = \frac{1}{2} h_n^\top H_{\theta^*} h_n +  h_n^\top \GG_n[\nabla L_{\theta^*}^{\eta^*}] + o_P(1).$$
At the same time, since $\P(\hat\eta\neq \eta^*) \to 0$, we have
$$n\E_n[L_{\theta^* + h_n/\sqrt{n}}^{\hat \eta} - L_{\theta^*}^{\hat \eta}] = n\E_n[L_{\theta^* + h_n/\sqrt{n}}^{\eta^*} - L_{\theta^*}^{\eta^*}] + o_P(1).$$
Putting everything together, we have shown
$$n\E_n[L_{\theta^* + h_n/\sqrt{n}}^{\hat \eta} - L_{\theta^*}^{\hat \eta}] = \frac{1}{2} h_n^\top H_{\theta^*} h_n +  h_n^\top \GG_n [\nabla L_{\theta^*}^{\eta^*}] + o_P(1).$$
The rest of the proof is standard. We apply the previous display with $h_n = \hat h_n := \sqrt{n}(\hat\theta^{\hat\eta} - \theta^*)$ (which is $O_P(1)$ by the consistency of $\hat\theta^{\eta^*}$; see \citep[Thm.~5.23]{van2000asymptotic}) and $h_n = \tilde h_n := -H_{\theta^*}^{-1}\GG_n[\nabla L_{\theta^*}^{\eta^*}]$: 
\begin{align*}
n\E_n[L_{\hat\theta^{\hat\eta}}^{\hat \eta} - L_{\theta^*}^{\hat \eta}] &= \frac{1}{2} \hat h_n^\top H_{\theta^*} \hat h_n +  \hat h_n^\top \GG_n [\nabla L_{\theta^*}^{\eta^*}] + o_P(1);\\
n\E_n[L_{\theta^* + \tilde h_n/\sqrt{n}}^{\hat \eta} - L_{\theta^*}^{\hat \eta}] &= \frac{1}{2} \tilde h_n^\top H_{\theta^*} \tilde h_n +  \tilde h_n^\top \GG_n [\nabla L_{\theta^*}^{\eta^*}] + o_P(1).
\end{align*}
By the definition of $\hat\theta^{\hat\eta}$, the left-hand side of the first equation is smaller than the left-hand side of the second equation. Therefore, the same must be true of the right-hand sides of the equations. If we take the difference between the equations and complete the square, we get
$$\frac 1 2 \left(\sqrt{n}(\hat\theta^{\hat\eta} - \theta^*) - \tilde h_n \right)^\top H_{\theta^*}\left(\sqrt{n}(\hat\theta^{\hat\eta} - \theta^*) - \tilde h_n\right) + o_P(1)\leq 0.$$
Since the Hessian $H_{\theta^*}$ is positive-definite, it must be the case that $\sqrt{n}(\hat\theta^{\hat\eta} - \theta^*) - \tilde h_n \stackrel{p}{\to} 0$. By the central limit theorem, $\tilde h_n = - H_{\theta^*}^{-1}\GG_n[\nabla L_{\theta^*}^{\eta^*}]$ converges to $\mathcal N(0,\Sigma_*)$ in distribution, where 
$$\Sigma_* = H_{\theta^*}^{-1} \Var\left(\nabla \ell_{\theta^*}(X,f(X))  + \left(\nabla \ell_{\theta^*}(X,Y) - \nabla \ell_{\theta^*}(X, f(X)) \right) \frac{\xi^{\eta^*}}{\pi_{\eta^*}(X)} \right)H_{\theta^*}^{-1}.$$
The final statement thus follows by Slutsky's theorem.

\subsection{Proof of Proposition \ref{prop:sequential_mean}}

We prove the result by an application of the martingale central limit theorem (see Theorem 8.2.4. in~\cite{durrett2019probability}).

Let $\bar \Delta_t$ denote the increments $\Delta_t$ with their mean subtracted out, i.e. $\bar \Delta_t = \Delta_t - \theta^*$. To apply the theorem, we first need to verify that the increments $\bar\Delta_t = \Delta_t - \theta^*$ are martingale increments; this follows because 
$$\E[\bar \Delta_t | \F_{t-1}] = \E[\bar \Delta_t | f_t,\pi_t] = \E[f_t(X_t)|f_t,\pi_t] + \E[Y_t - f_t(X_t)|f_t,\pi_t] \E\left[\frac{\xi_t}{\pi_t(X_t)}|f_t,\pi_t\right] - \theta^* = 0,$$
together with the fact that $\bar \Delta_t \in \F_t$.

The martingale central limit theorem is now applicable given two regularity conditions. The first is that $\frac 1 n \sum_{t=1}^n \sigma_t^2$ converges in probability, which holds by assumption.
The second condition is the so-called Lindeberg condition, stated below.

\begin{assumption}
\label{ass:lindeberg_mean}
Let $\bar \Delta_t = \Delta_t - \theta^*$. We say that $\Delta_t$ satisfy the Lindeberg condition if for all $\epsilon >0$,
$$\frac 1 n \sum_{t=1}^n \E[\bar \Delta_t^2 \one\{|\bar \Delta_t| > \epsilon \sqrt{n}\} | \F_{t-1}]\stackrel{p}{\to} 0.$$
\end{assumption}
\noindent Since this condition holds by assumption, we can apply the central limit theorem to conclude $\sqrt{n}(\hat\theta^{\vv{\pi}} - \theta^*) = \frac{1}{\sqrt{n}}\sum_{t=1}^n \bar \Delta_t \cd \cN(0,  \sigma^2_*).$

\subsection{Proof of Theorem \ref{thm:sequential_general}}

We follow a similar approach as in the proof of Theorem \ref{thm:m_est_batch}, which is in turn similar to the classical argument for M-estimation \citep[Thm.~5.23]{van2000asymptotic}. The main difference from the classical proof is that the empirical loss \(L^{\vv{\pi}}(\theta)\) is built from martingale, rather than i.i.d. increments. We explain the differences relative to the proof of Theorem \ref{thm:m_est_batch}.

We define $L_{\theta,i}
=
\ell_\theta(X_i,f_i(X_i))
+
\Big(\ell_\theta(X_i,Y_i)-\ell_\theta(X_i,f_i(X_i))\Big)\frac{\xi_i}{\pi_i(X_i)}$,
and \(\nabla L_{\theta,i}\), \(\nabla^2 L_{\theta,i}\) are defined analogously. We again use the notation
\(\GG_n[g(L_\theta)]\), \(\E_n[g(L_\theta)]\), \(\GG_n[g(\nabla L_\theta)]\), \(\E_n[g(\nabla L_\theta)]\), etc.

As in the classical argument, for any \(h_n = O_P(1)\), we have $\GG_n\!\left[\sqrt{n}\big(L_{\theta^* + h_n/\sqrt{n}}-L_{\theta^*}\big)-h_n^\top \nabla L_{\theta^*}\right]
\stackrel{p}{\to} 0$. This can be concluded from the martingale central limit theorem. To see this, define for any $h\in\R^d$:
$G_n(h)
:=
\GG_n\!\left[\sqrt{n}\big(L_{\theta^*+h/\sqrt{n}}-L_{\theta^*}\big)-h^\top \nabla L_{\theta^*}\right].$
We argue that $\sup_{\|h\|\le C} |G_n(h)| \stackrel{p}{\to} 0$ for every fixed $C$, which immediately implies $G_n(h_n)\stackrel{p}{\to}0$ for every $h_n=O_P(1)$.
By a second-order Taylor expansion around $\theta^*$, for each fixed $h$,
\[
\sqrt{n}\big(L_{\theta^*+h/\sqrt{n},i}-L_{\theta^*,i}\big)
=
h^\top \nabla L_{\theta^*,i}
+
\frac{1}{2\sqrt{n}}\, h^\top \nabla^2 L_{\theta^*,i} h
+
r_{n,i}(h),
\]
where the remainder satisfies $|r_{n,i}(h)|
\le
\frac{\|h\|^2}{2\sqrt{n}}
\sup_{\|\theta-\theta^*\|\le \|h\|/\sqrt{n}}
\big\|\nabla^2 L_{\theta,i}-\nabla^2 L_{\theta^*,i}\big\|.$
Summing and centering gives
\[
G_n(h)
=
\frac12 h^\top \GG_n[\nabla^2 L_{\theta^*}] h + \GG_n\![\sum_{i=1}^n r_{n,i}(h)].
\]

Since \(\nabla^2 L_{\theta^*,i}\) are martingale increments after centering, the martingale CLT implies $\GG_n[\nabla^2 L_{\theta^*}] \stackrel{p}{\to} 0.$
For the remainder term, we use the fact that the Hessian is locally Lipschitz in a neighborhood of $\theta^*$ to conclude that, for any $C$, $\sup_{\|h\|\le C} |\GG_n\![\sum_{i=1}^n r_{n,i}(h)]| \stackrel{p}{\to} 0.$
Combining the two steps yields $
\sup_{\|h\|\le C} |G_n(h)| \stackrel{p}{\to} 0$ for every fixed $C<\infty$, which in turn implies the statement for all $h_n=O_P(1)$.

The following steps are the same as in the proof of Theorem \ref{thm:m_est_batch}; we conclude that $\sqrt{n}(\hat\theta^{\vv{\pi}} - \theta^*) - \tilde h_n \cp 0$, where $\tilde h_n = - H_{\theta^*}^{-1}\GG_n[\nabla L_{\theta^*}]$. Finally, we argue that $\tilde h_n$ converges to $\mathcal N(0,\Sigma_*)$ in distribution. To see this, first note that all one-dimensional projections $v^\top \tilde h_n$ converge to $v^\top Z$, $Z\sim \mathcal N(0,\Sigma_*)$, by the martingale central limit theorem, which is applicable because the Lindeberg condition holds by assumption (see below for statement) and the variance process $V_{\theta^*,n}$ converges to $V_*$. Once we have the convergence of all one-dimensional projections, convergence of $\tilde h_n$ follows by the Cram\'er-Wold theorem.

\begin{assumption}
\label{ass:lindeberg_general}
We say that the increments satisfy the Lindeberg condition if, for all $v\in \mathcal S^{d-1}$ and $\epsilon >0$,
$$\frac 1 n \sum_{t=1}^n \E[(v^\top \nabla L_{\theta^*, t})^2 \one\{|v^\top \nabla L_{\theta^*,t}| > \epsilon \sqrt{n}\} | \F_{t-1}]\stackrel{p}{\to} 0.$$
\end{assumption}

\section{Experimental details}
\label{app:exps}

In all our experiments, we have a labeled dataset of $n$ examples. We treat the solution on the full dataset as the ground-truth $\theta^*$ for the purpose of evaluating coverage. In each trial, the underlying data points $(X_i, Y_i)$ are fixed and the randomness comes from the labeling decisions $\xi_i$. In the sequential experiments, we additionally randomly permute the data points at the beginning of each trial. The experiments in the batch setting average the results over $1000$ trials and the experiments in the sequential setting average the results over $100$ trials.
The Pew dataset is available at \cite{atp79}; the census dataset is available through Folktables~\cite{ding2021retiring}; the Alphafold dataset is available at \cite{ppidata}.

As discussed in Section \ref{sec:practical_rules}, to avoid values of $\pi(x)$ that are close to zero we mix the ``standard'' sampling rule based on the uncertainty $u(x)$ with a uniform rule $\pi^{\mathrm{unif}} = \frac{n_b}{n}$ according to a parameter $\tau\in(0,1)$. In post-election survey research, we have training data for the prediction model and we use the same data to select $\tau$ so as to minimize an empirical approximation of the variance $\Var(\hat\theta^{\pi^{(\tau)}})$, as in Eq.~\eqref{eqn:tau-tuning}. In the AlphaFold example and both problems with model fine-tuning we set $\tau = 0.5$ for simplicity.
In the census example, the trained predictor of model error $e(x)$ rarely gives very small values, and so we set $\tau = 0.001$.

In each experiment, we vary $n_b$ over a grid of uniformly spaced values. We take $20$ grid values for the batch experiments and $10$ grid values for the sequential experiments. The plots of interval width and coverage linearly interpolate between the respective values obtained at the grid points. There linearly interpolated values are used to produce the plots of budget save: for all values of $n_b$ from the grid, we look for $n_b'$ such that the (linearly interpolated) width of active inference at sample size $n_b'$ matches the interval width of classical (resp. uniform) inference at sample size $n_b$. 
For the leftmost plot in Figures \ref{fig:pew79_batch}-\ref{fig:alphafold} and Figures \ref{fig:pew79_finetuning}-\ref{fig:census_finetuning}, we uniformly sample five trials for a fixed $n_b$ and show the intervals for all methods in those same five trials. We arbitrarily select $n_b$ to be the fourth largest value in the grid of budget sizes for all experiments.

\section{Non-asymptotic results}
\label{sec:nonasymptotic}

While our results focus on asymptotic confidence intervals based on the central limit theorem, some of them---in particular, those for mean estimation---have direct non-asymptotic and time-uniform analogues.

We explain this extension for the sequential algorithm, as it subsumes the extension for the batch setting. Let $\Delta_t = f_t(X_t) + (Y_t - f_t(X_t)) \frac{\xi_t}{\pi_t(X_t)}$. As explained in Section \ref{sec:sequential}, $\Delta_t$ have a common conditional mean: $\E[\Delta_t|\Delta_1,\dots,\Delta_{t-1}] = \theta^*$. Moreover, if $Y_t$ and $f_t(X_t)$ are almost surely bounded, and $\pi_t(X_t)$ is almost surely bounded from below, then $\Delta_t$ are bounded as well. (Given that we construct $\pi_t$ by ``$\tau$-mixing'' it with a uniform rule, as explained in Section \ref{sec:practical_rules}, in our applications $\pi_t(X_t)$ is always bounded from below since
$\pi_t(x) \geq \tau \frac{n_b}{n}$.) Therefore, given that we have bounded observations (with a known bound) having a common conditional mean, we can apply the recent betting-based methods \cite{waudby2020estimating, orabona2023tight} for constructing non-asymptotic confidence intervals and time-uniform confidence sequences satisfying $\P(\theta^* \in C_t, \forall t) \geq 1-\alpha$.

\begin{figure}[h]
\centering
\includegraphics[width=\textwidth]{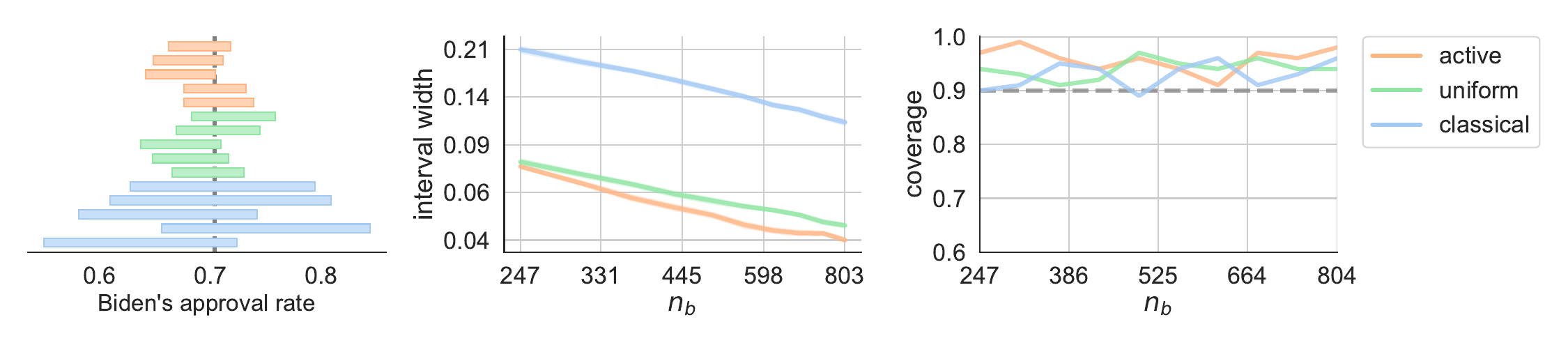}
\includegraphics[width=\textwidth]{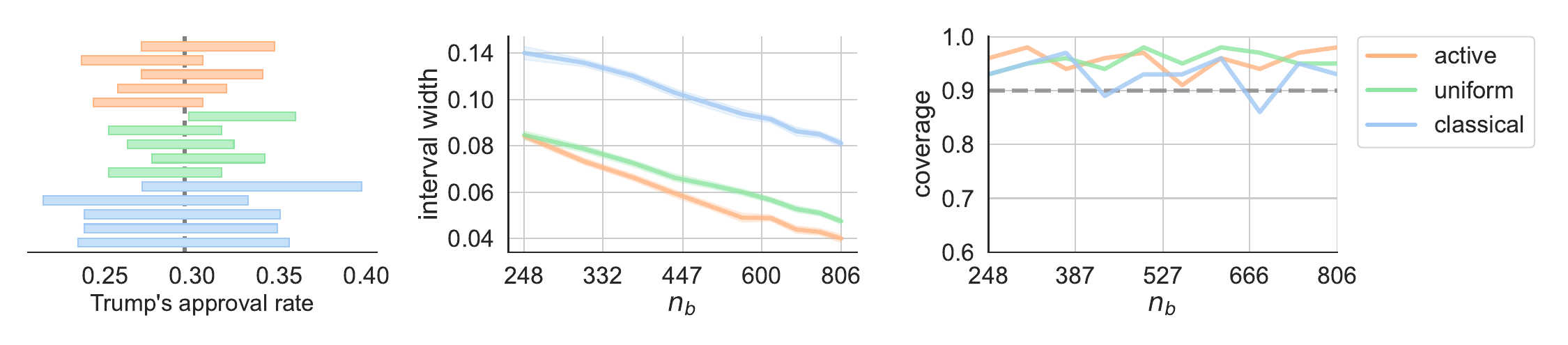}
\caption{\textbf{Non-asymptotic experiments.} Example intervals in five randomly chosen trials (left), average confidence interval width (middle), and coverage (right) in post-election survey research with non-asymptotic confidence intervals.}
\label{fig:nonasymp}
\end{figure}

We demonstrate the non-asymptotic extension in the problem of post-election survey analysis from Section~\ref{sec:election}. Figure \ref{fig:nonasymp} provides a non-asymptotic analogue of the corresponding batch results from Figure \ref{fig:pew79_batch}, applying the method from Theorem~3 of \citet{waudby2020estimating} to form a non-asymptotic confidence interval.
Qualitatively we observe a similar comparison as before---active inference outperforms both uniform sampling and classical inference---though the methods naturally overcover as a result of using non-asymptotic intervals that do not have exact coverage.

\end{document}